\pdfoutput=1
\documentclass[11pt]{article}

\usepackage[preprint]{acl}

\usepackage{times}
\usepackage{latexsym}

\usepackage[T1]{fontenc}
\usepackage[utf8]{inputenc}

\usepackage{microtype}

\usepackage{inconsolata}

\usepackage{graphicx}

\usepackage{multirow} % Required for multirows
\usepackage{tcolorbox} % For tcolorbox environment
\usepackage{amsmath,amssymb,amsfonts}%
\usepackage{wasysym}
\usepackage{amsthm}%
\usepackage{mathrsfs}%
\usepackage{xcolor}%
\usepackage{textcomp}%
\usepackage{manyfoot}%
\usepackage{booktabs}%
\usepackage{algorithm}%
\usepackage{algorithmicx}%
\usepackage{algpseudocode}%
\usepackage{listings}%
\usepackage{array}
\usepackage{makecell}

\usepackage{enumitem}
\usepackage{pifont} % For checkmark and cross
\usepackage{booktabs}
\usepackage{subfig}
\usepackage{hyperref}
\usepackage{ragged2e}
\usepackage{booktabs}
\usepackage{multirow}
\usepackage{pifont}
\usepackage{xcolor}
\usepackage{cuted}   % 用于创建跨栏的 strip 环境
\usepackage{capt-of} % 用于在非浮动环境中添加 caption
\usepackage{pifont} % 用于 \cmark, \xmark
\usepackage[table]{xcolor}

\newcommand{\cmark}{\ding{51}}%
\newcommand{\xmark}{\ding{55}}%
\definecolor{mygray}{gray}{0.9}

\title{Benchmarking Egocentric Clinical Intent Understanding Capability \\ for Medical Multimodal Large Language Models} 

\author{xxx
    }

\author{
    Shaonan Liu$^{1}$\thanks {Equal contribution.}, 
    Guo Yu$^{1}$\footnotemark[1], 
    Xiaoling Luo$^{1}$, \\
    \textbf{Shiyi Zheng}$^{1}$, 
    \textbf{Wenting Chen}$^{2}$\thanks{Corresponding Author.}, 
    \textbf{Jie Liu}$^{3}$\footnotemark[2], 
    \textbf{Linlin Shen}$^{1}$\footnotemark[2]\\
    $^{1}$Shenzhen University, 
    $^{2}$Stanford University, 
    $^{3}$City University of Hong Kong
}
\begin{document}
\maketitle

\begin{abstract}
Medical Multimodal Large Language Models (Med-MLLMs) require egocentric clinical intent understanding for real-world deployment, yet existing benchmarks fail to evaluate this critical capability. To address these challenges, we introduce \textbf{MedGaze-Bench}, the first benchmark leveraging clinician gaze as a Cognitive Cursor to assess intent understanding across surgery, emergency simulation, and diagnostic interpretation. Our benchmark addresses three fundamental challenges: visual homogeneity of anatomical structures, strict temporal-causal dependencies in clinical workflows, and implicit adherence to safety protocols. We propose a \textbf{Three-Dimensional Clinical Intent Framework} evaluating: (1) \textit{Spatial Intent}—discriminating precise targets amid visual noise, (2) \textit{Temporal Intent}—inferring causal rationale through retrospective and prospective reasoning, and (3) \textit{Standard Intent}—verifying protocol compliance through safety checks. Beyond accuracy metrics, we introduce Trap QA mechanisms to stress-test clinical reliability by penalizing hallucinations and cognitive sycophancy. Experiments reveal current MLLMs struggle with egocentric intent due to over-reliance on global features, leading to fabricated observations and uncritical acceptance of invalid instructions. %MedGaze-Bench provides a rigorous framework for advancing clinically safe, context-aware Med-MLLMs.

\end{abstract}

\section{Introduction}
%%%
%%我们的数据集有：开放式手术 胎儿臀围分娩 医生阅片诊断

%%%%%%
%%Spatial Intent Understanding：(1) Absolute Localization (2) Relative Localization

%%Temporal Intent Understanding：(1) Prospective Anticipation (2) Retrospective Attribution

%%Standard Intent Understanding：(1) Discrete Safety Verification (2) Continuous Safety Vigilance

%%%Intent Cognition

The capabilities of Medical Multimodal Large Language Models (Med-MLLMs) have evolved substantially, from report generation to complex clinical reasoning \cite{li2023llava,yu2025finemedlm} and sequential decision-making in multi-turn dialogues \cite{liu2024medchain,li2024agent}. To further advance toward fully AI Doctor assistant, these models must perceive, reason, and interact from an egocentric perspective that mirrors real-world clinical workflows. 

However, existing benchmarks fall short in evaluating such capabilities. While datasets like EgoSurgery~\cite{fujii2024egosurgery} and POV-Surgery~\cite{wang2023pov} support specific visual tasks, they do not evaluate Med-MLLMs' intent understanding capacity, i.e., the underlying purpose and rationale behind clinician actions across diverse scenarios ranging from surgical interventions to diagnostic interpretations. Therefore, systematic benchmarks are needed to evaluate Med-MLLMs' capabilities in understanding clinician actions.

%% origin
% Medical Multimodal Large Language Models (Med-MLLMs) have evolved rapidly, transitioning from initial report generation to mastering complex clinical reasoning~\cite{li2023llava,yu2025finemedlm}. Recent advances have even demonstrated their potential for sequential decision-making in multi-turn dialogues~\cite{liu2024medchain,li2024agent}. To advance towards fully autonomous "AI Doctor" agents, it is imperative for these models to transcend passive observation and learn to perceive, reason, and interact from an Egocentric View, mirroring the immersive and dynamic nature of real-world clinical procedures. However, despite this immense potential, there is a distinct scarcity of comprehensive benchmarks tailored for this direction. While prior works like EgoSurgery~\cite{fujii2024egosurgery} and POV-Surgery~\cite{wang2023pov} have laid the data foundation for specific visual tasks, they fall short of evaluating the intent-level cognition of Med-MLLMs. Current evaluations often miss the holistic capabilities required across varied clinical contexts—ranging from surgical interventions to diagnostic interpretations. The absence of a systematic evaluation standard to understand why a clinician acts, rather than just what they do, remains a significant bottleneck.

\begin{figure}[t]
\centering
	{\includegraphics[width=1\linewidth]%{Fig/Fig1_Big_teaser_1.png}}
    {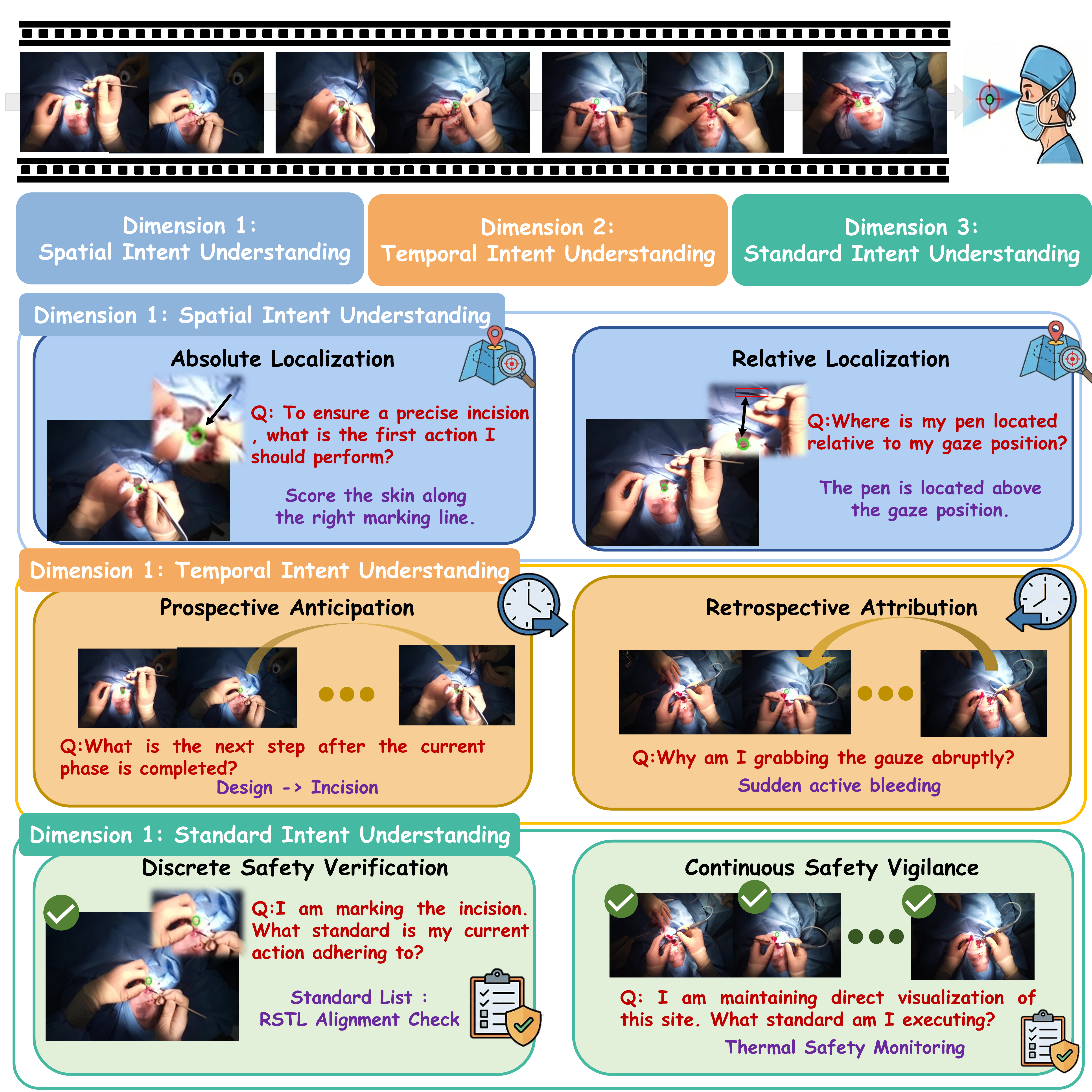}}
	% \caption{Overview of the MedGaze-Bench and the Three-Dimensional Clinical Intent Framework
    % }
	\caption{\textbf{MedGaze-Bench} with three-dimensional clinical intent understanding capability evaluation from \textbf{\textit{spatial}}, \textbf{\textit{temporal}} to \textbf{\textit{medical standard}}.}
	\label{fig:MedGaze_teaser}

\end{figure}

To construct an effective evaluation framework, we should first analyze three intrinsic challenges in egocentric medical scenarios that hinder intent understanding:

\noindent\textbf{(1) High Visual Homogeneity and Ambiguity}: In operative fields and diagnostic displays, targets often lack distinct visual boundaries. Nerves and vessels exhibit similar textures in surgery~\cite{maier2022surgical}, while subtle pathological signs blend with healthy tissue in diagnostics~\cite{yan2018deeplesion}. Global features alone cannot distinguish the clinician's immediate focus from visually dominant surroundings. Hence, we should define both global localization and relative localization task.

\noindent\textbf{(2) Strict Temporal and Causal Dependency}: Clinical actions are rarely isolated; they are linked by inherent causal necessities~\cite{liu2023multilevel}. Achieving hemostasis before closure in surgery, or localizing lesions before characterization in diagnostics~\cite{kundel1978visual}, exemplifies mandatory sequential dependencies. Models must understand that violating this temporal order is not merely a stylistic error, but a fundamental failure in clinical reasoning.

\noindent\textbf{(3) Implicit Standardization based on Guidelines}: Clinical actions follow Standard Operating Procedures (SOPs) with latent safety logic. In vaginal breech delivery, specific hand movements (e.g., Lovset or Mauriceau maneuvers) are mandated by obstetric guidelines to prevent complications~\cite{latifzadeh2025assessing,becker2025royal}, not arbitrary choices. Failing to grasp the latent safety logic behind these standardized motions prevents models from truly comprehending the procedure.

% MedGaze-Bench: Benchmarking Egocentric Clinical Intent
% Understanding Capability for Medical Multimodal Large Language Models
To fulfill these requirement, we introduce \textbf{MedGaze-Bench}, the first medical benchmark designed to evaluate MLLMs' capabilities in egocentric clinical intent understanding. It is constructed from three distinct clinical scenarios: unstructured real-world open surgery (20 procedures by 8 surgeons), standardized emergency simulation (breech delivery by 5 obstetricians), and fine-grained diagnostic cognition (Chest X-ray and Mammography interpretation). Despite their heterogeneity, we identify a unifying cognitive thread: the \textbf{Gaze}. We conceptualize Gaze as a ``\textbf{Cognitive Cursor}''—a dynamic proxy bridging raw visual stimuli and high-level clinical reasoning. A clinician's gaze explicitly indicates their implicit reasoning: it filters visual noise (Spatial), reveals procedural anticipation (Temporal), and verifies safety protocols (Standard).

\begin{table*}[t]
\centering
\caption{Comparison with representative egocentric and medical benchmarks.}%\textbf{Gaze Signals} with medical intent reasoning, uniquely covering \textbf{SOPs Compliance} verification.}
%\textbf{MedGaze-Bench} is the first to bridge high-fidelity \textbf{Gaze Signals} with medical intent reasoning, uniquely covering \textbf{SOPs Compliance} verification.}
\label{tab:benchmark_comparison}

% Optional: Use \small or \footnotesize instead of resizebox for better readability
\resizebox{\textwidth}{!}{%
% Change: Removed the last "|c" in the column definition
\begin{tabular}{l|cc|c|c|ccc}
\toprule
\multirow{2}{*}{\textbf{Benchmark}} & \multicolumn{2}{c|}{\textbf{Data Scale}} & \textbf{View} & \textbf{Gaze} & \multicolumn{3}{c}{\textbf{Intent Reasoning Capabilities}} \\ % Change: Removed Strategy Header
 & \textbf{Videos} & \textbf{QA Pairs} & \textbf{Persp.} & \textbf{Source} & \textbf{Spatial} & \textbf{Temporal} & \textbf{Standard} \\ \midrule % Change: Removed Prompts Header

% --- General Domain ---
% Change: Changed multicolumn span from 9 to 8
\multicolumn{8}{l}{\textit{\textbf{General Domain}}} \\
QaEgo4D \cite{barmann2022did} & 166 & 1,854 & Ego & \xmark$^{\dagger}$ & \cmark & \cmark & \xmark \\
EgoMemoria \cite{ye2024mm} &  629 & 7,026 & Ego & \xmark$^{\dagger}$ & \cmark & \cmark & \xmark \\

ECBench \cite{dang2025ecbench} & 386 & 4,324 & Ego & \xmark$^{\dagger}$ & \cmark & \cmark & \xmark \\
EOC-Bench \cite{yuan2025eoc} & 656 & 3,277 & Ego & \xmark & \cmark & \cmark & \xmark \\
EgoTextVQA \cite{zhou2025egotextvqa} & 1,507 & 7,064 & Ego & \xmark & \cmark & \cmark & \xmark \\
EgoGazeVQA \cite{peng2025eye} & 913 & 1,757 & Ego & \cmark  & \cmark & \cmark & \xmark \\
\midrule

% --- Medical Domain ---
% Change: Changed multicolumn span from 9 to 8
\multicolumn{8}{l}{\textit{\textbf{Medical Domain}}} \\

Cholec80 \cite{maier2022surgical} & 80 & \textasciitilde43,182 & Non-ego & \xmark & \cmark & \xmark & \xmark \\
%EndoBench \cite{liu2025endobench} & 6,832 imgs & 6,832 & Non-ego & \xmark & \cmark & \xmark & \xmark \\
EndoBench \cite{liu2025endobench} & 6,832$^{\ddagger}$ & 6,832 & Non-ego & \xmark & \cmark & \xmark & \xmark \\
%SSG-VQA \cite{ecosurgery2024} & 15 & - & Non-Ego & \xmark & \xmark & \cmark & \xmark \\
%SurgPub-Video \cite{li2025surgpub} & 10,926 &48,520 & Non-ego & \xmark & \cmark & \cmark & \xmark \\

EgoSurgery \cite{fujii2024egosurgery} & 571 &- & Ego & \cmark & \cmark & \xmark & \xmark \\

EgoExOR \cite{ozsoy2025egoexor} & 41 &- & Ego & \cmark & \cmark & \xmark & \xmark \\
\midrule

% --- Ours ---
\rowcolor{mygray}
\textbf{MedGaze-Bench} & \textbf{775} & \textbf{4,491} & \textbf{Ego} & \textbf{\cmark} & \textbf{\cmark} & \textbf{\cmark} & \textbf{\cmark} \\ \bottomrule
\end{tabular}%
}

\vspace{5pt}
\footnotesize{
\textbf{View Persp.}: Ego=Egocentric, Endo=Endoscopic.
\textbf{Gaze Source}: EgoSurgery uses IMU-based head motion; Ours uses Eye-Tracking.
$^{\dagger}$: Ego4D contains gaze subset, unused in standard QA.
% Change: Added explanation for \ddagger
$^{\ddagger}$: Dataset consists of static images.
}
\end{table*}

We establish the \textbf{Three-Dimensional Clinical Intent Framework} evaluating how MLLMs synthesize visual stimuli into actionable reasoning from {\textit{spatial}}, {\textit{temporal}}, and {\textit{medical standard}} perspectives (Figure~\ref{fig:MedGaze_teaser}). \textit{First}, \textbf{Spatial Intent Understanding} addresses visual ambiguity (the "\textbf{\textit{Where}}") through Discriminative Grounding, requiring models to filter visual noise and identify precise anatomical targets via Absolute Localization while decoding surrounding spatial logic via Relative Localization. \textit{Moreover}, \textbf{Temporal Intent Understanding} tackles causal dependency (the "\textbf{\textit{Why}}") through Causal Rationale, evaluating whether models perform Retrospective Attribution to deduce prerequisite conditions justifying current actions and Prospective Anticipation to forecast operational goals driving next steps, transcending mere chronological sequence. \textit{Furthermore}, \textbf{Standard Intent Understanding} captures SOP adherence (the "\textbf{\textit{How}}") through Protocol Alignment, decomposed into Discrete Safety Verification for momentary checking of critical landmarks and Continuous Safety Vigilance for sustained monitoring of vulnerable non-target areas.

%% origin 
% \textit{1) \textbf{Spatial Intent Understanding}} (The ``Where'' of Attention): Addressing visual ambiguity, this dimension evaluates Discriminative Grounding. Instead of passive detection, models must actively filter high-frequency visual noise to lock onto the precise anatomical target (Absolute Localization) and decode its surrounding spatial logic (Relative Localization), mirroring a surgeon’s tunnel vision during critical maneuvers.
% \textit{2) \textbf{Temporal Intent Understanding}} (The Why'' of Reasoning): Addressing strict causal dependency, this dimension measures Causal Rationale. It evaluates whether the model utilizes current gaze to deduce the prerequisite conditions that justify the present action (Retrospective Attribution) and forecast the operational goals that drive the next step (Prospective Anticipation), fundamentally transcending the mere When'' of chronological sequence.
% \textit{3) \textbf{Standard Intent Understanding}} (The ``How'' of Safety): Capturing the adherence to SOPs, this dimension measures Protocol Alignment. We decompose this intent into two complementary safety mechanisms: (Discrete Safety Verification), which necessitates the momentary checking of critical landmarks before irreversible actions; and (Continuous Safety Vigilance), which demands sustained monitoring of vulnerable non-target areas to prevent collateral injury.

We conduct extensive evaluation of 9 MLLMs on MedGaze-Bench, revealing that current MLLMs frequently fail to interpret clinical intent from egocentric videos. Several obtained insights offer potential direction of Med-MLLMs.
Our contributions are as follows:

%The First Gaze-Centric Medical Intent Benchmark:
\begin{itemize} \item  We introduce MedGaze-Bench, the first benchmark utilizing clinician gaze as a ``Cognitive Cursor'' to bridge the critical gap between passive egocentric perception and active clinical reasoning.
%A Systematic Three-Dimensional Intent Framework: 
\item We propose a unified framework evaluating \textit{Spatial}, \textit{Temporal}, and \textit{Standard Intent Understanding}. This structure systematically quantifies how models handle visual ambiguity, causal dependency, and rigorous safety protocols.

%A Hierarchical Competency-Reliability Protocol: 
\item Beyond standard accuracy, we design a dual-level evaluation strategy featuring a novel ``Trap QA'' mechanism. This explicitly stress-tests clinical reliability, strictly penalizing models for perceptual hallucinations and cognitive sycophancy.
\end{itemize}

\begin{figure*}[t]
\centering

	{\includegraphics[width=1\linewidth]{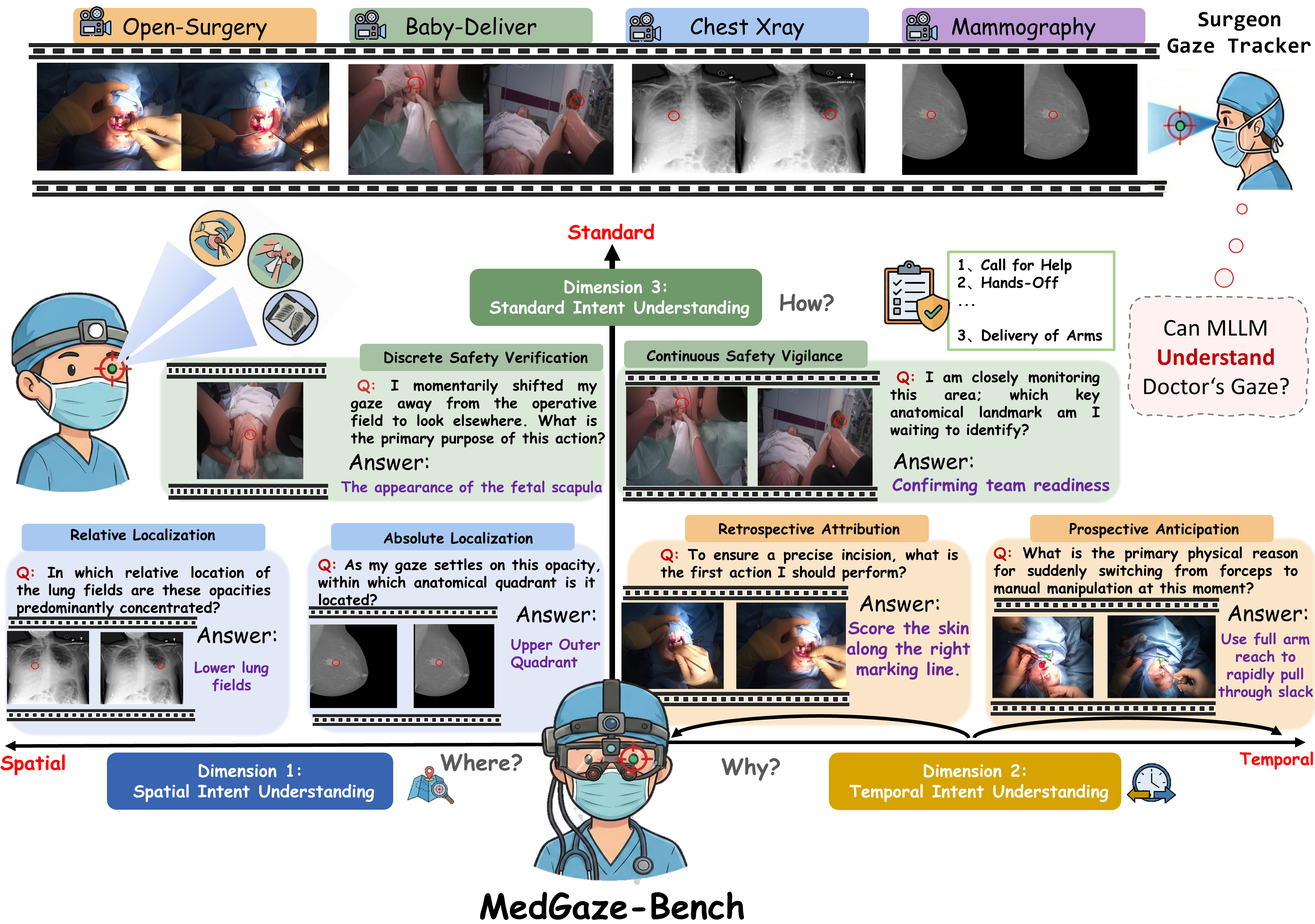}}

	\caption{\textbf{MedGaze-Bench,} a three-dimensional evaluation system for clinical intent understanding based on gaze tracking across four medical scenarios. It assesses \textbf{spatial intent} (Where?), \textbf{temporal intent} (Why?), and \textbf{standard intent understanding} (How?), using gaze as a cognitive proxy to evaluate MLLMs' ability to bridge visual perception and clinical decision-making.
}
	\label{fig:main_framework}
\end{figure*}

%多模态 Medical
%\subsection{Medical Video Understanding \& QA Benchmarks}

\section{Related Work}

\subsection{General VideoQA Benchmarks} Video Question Answering (VideoQA) has evolved significantly, transitioning from identifying atomic visual patterns to requiring high-level cognitive synthesis. While foundational benchmarks established the groundwork for cross-modal alignment through tasks like action recognition~\cite{caba2015activitynet,deng2023large} and video captioning~\cite{takahashi2024abstractive}, they often treated videos as sequences of static frames, neglecting underlying causal dynamics. To address this limitation, recent benchmarks such as MVBench~\cite{li2024mvbench} and Video-MME~\cite{fu2025video} have expanded the scope to include temporal grounding and long-context understanding across diverse daily scenarios. However, despite the shift towards more complex reasoning~\cite{hu2025video}, current general VideoQA benchmarks remain predominantly constrained by a third-person "spectator" perspective. This observation angle creates an inherent information asymmetry, lacking the egocentric immersion and fine-grained cognitive signals—specifically gaze—required to decode the implicit intent of professional practitioners in high-stakes environments.

\subsection{Medical VideoQA Benchmarks} 
The rapid proliferation of Medical Multimodal Large Language Models (Med-MLLMs) has spurred the development of diverse evaluation suites. While broad-spectrum benchmarks like GMAI-MMBench~\cite{ye2024gmai} and MediConfusion~\cite{sepehri2024mediconfusion} have established baselines for modality coverage and discriminative robustness, significant efforts have also been directed towards dynamic surgical environments. Specialized benchmarks, including SurgicalVQA~\cite{seenivasan2022surgical} and EndoBench~\cite{liu2025endobench}, have advanced the field by aggregating datasets to evaluate geometric localization and procedural phase analysis. Nevertheless, these existing works remain predominantly constrained by a post-hoc "observer bias". They rely on external annotations that describe \textit{what} is happening (e.g., tool presence or phase labels) but fail to capture \textit{why} the clinician acted at that specific moment, effectively detaching visual input from the clinician's active cognitive process. MedGaze-Bench addresses this critical gap by introducing an egocentric, gaze-centric paradigm, shifting the evaluation from passive event observation to active intent understanding. Table~\ref{tab:benchmark_comparison} provides a systematic comparison, highlighting that our MedGaze-Bench is the first to uniquely integrate egocentric vision, authentic gaze signals, and SOP-based intent verification.

\subsection{Gaze in Clinical Cognition}

In the medical domain, gaze extends beyond a mere point of visual fixation; it acts as a physical manifestation of systematic reasoning, rooted in the "Eye-Mind Hypothesis"~\cite{anderson2004eye}. Building on this cognitive link, recent approaches have integrated gaze signals to enhance model robustness. For instance, Wang et al.~\cite{wang2025improving} aligned visual representations with human gaze during self-supervised pre-training to prioritize clinical features, while Ma et al.~\cite{ma2023eye} employed gaze guidance to rectify "shortcut learning" in Vision Transformers. Parallel to these diagnostic advances, the surgical domain has transitioned towards first-person video analysis to capture the immersive nature of procedures. Benchmarks like EgoSurgery~\cite{fujii2024egosurgery} have pioneered fine-grained action recognition from the surgeon's perspective. However, these works typically focus on recognizing \textit{visible} hand-object interactions ("What is happening"), ignoring the \textit{implicit} attention signals ("Where the surgeon is planning"). MedGaze-Bench addresses this disparity by unifying gaze-driven cognition with egocentric video, shifting the evaluation paradigm from passive observation to active intent understanding.

%1)早期的医学 VQA 测评主要关注静态图像VQA-Med 和 PathVQA。
%2)随着发展手术视频理解...
%3)但是视角局限，任务基础
%4)现有的医学场景下缺乏对医生第一视角下的意图评估
%5)我们提出了首个第一视角下的临床医生凝视意图

%Egocentric General 
%Gaze in medicne
%\subsection{Egocentric Gaze and Actions in Medicine}
%1)第一人称视角（Egocentric View）在通用领域（如 Ego4D, EPIC-KITCHENS）已不仅限于动作识别，还涉及意图预测。
%2)在医学领域，EgoSurgery [cite] 填补了开放式手术数据的空白，并利用头部运动（Head motion）作为注意力的代理来辅助动作识别

%3)具体的眼动追踪（Eye-tracking）在医学中长期被用作临床熟练度（Clinical Proficiency）的生理指标。研究表明，无论是外科医生还是放射科医生，专家的注视模式都反映了更高效的视觉搜索策略和认知负荷管理 [cite]。
%4)然而，现有的研究主要将 Gaze 视为一种事后分析工具（Retrospective Analysis Tool）（用来评价人）。鲜有工作将其转化为实时的输入信号（Real-time Input Signal）来增强 AI 对临床意图的理解。
%我们提出将这种“专家视线”作为提示，弥合了像素感知与临床决策逻辑之间的鸿沟。

\section{MedGaze-Bench}%Ours Benchmark}

\begin{figure}[t]
\centering
	{\includegraphics[width=1\linewidth]{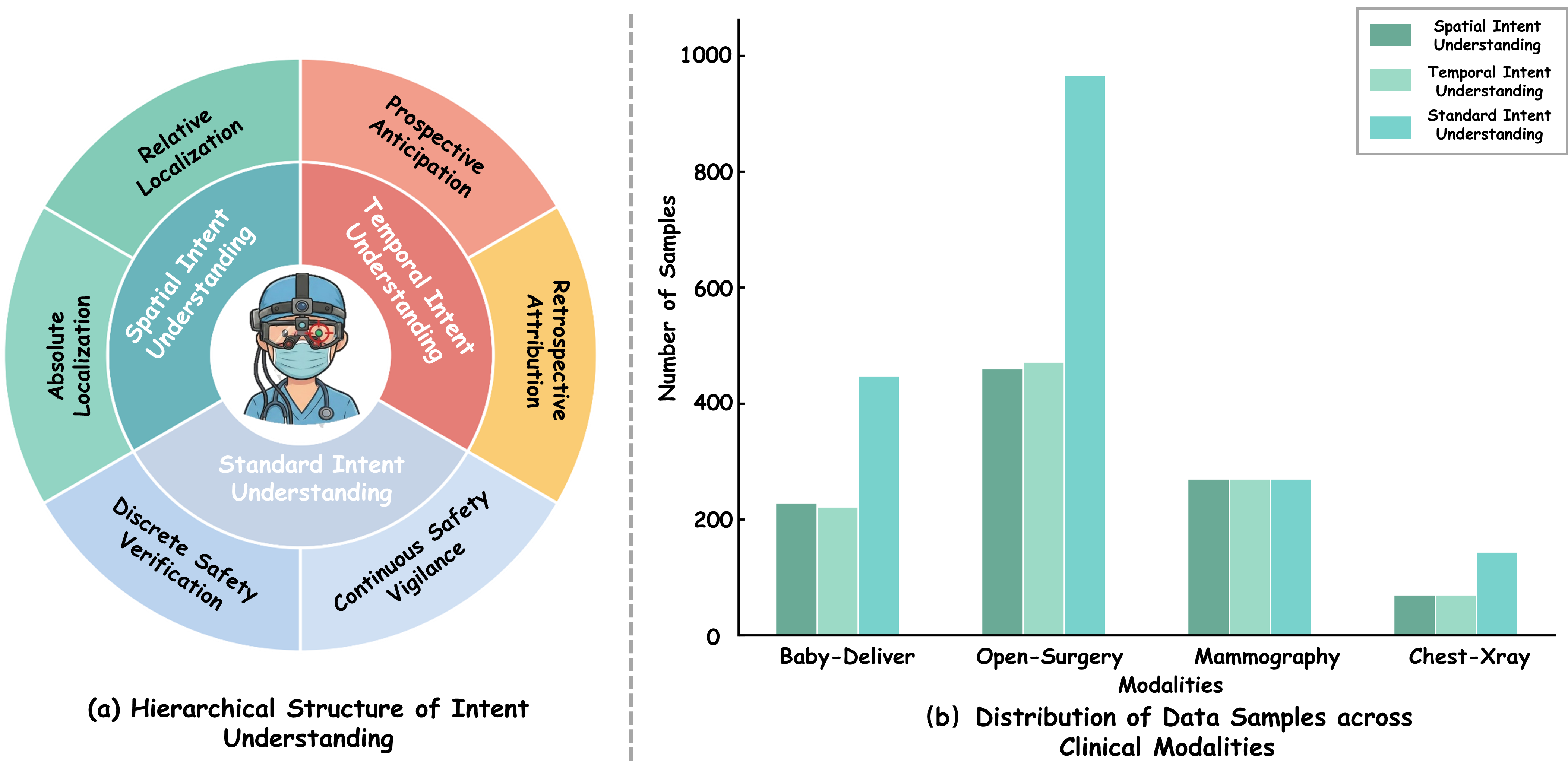}}
	% \caption{\textbf{Overview of the MedGaze-Bench Framework and Statistics.} (a) The hierarchical taxonomy decomposing the three primary intents into six fine-grained sub-capabilities. (b) The data distribution across four clinical scenarios, demonstrating the benchmark's diversity and coverage.} 
	\caption{(a) MedGaze-Bench categories. (b) Data distribution across 4 clinical scenarios.} \label{fig:framework_and_stats}
\end{figure}

\noindent\textbf{Overview.} We introduce \textbf{MedGaze-Bench}, a benchmark for evaluating clinical intent understanding across three scenarios: Open Surgery, Emergency Simulation, and Diagnostic Radiology (Figure~\ref{fig:main_framework}). The benchmark features a Three-Dimensional Clinical Intent Framework with six fine-grained sub-capabilities that reflect hierarchical expert reasoning, comprising 4,491 clinically validated samples. In Figure~\ref{fig:framework_and_stats}, the data distribution mirrors medical task dynamics: Temporal Intent Understanding (2,028 samples) dominates due to dense causal dependencies, while Standard (1,234) and Spatial Intent Understanding (1,229) are balanced. The benchmark incorporates a rigorous Clinical Evaluation Protocol with a novel ``Trap QA'' component (600 adversarial samples) that explicitly targets Perceptual and Cognitive Hallucinations to assess Clinical Reliability against visual fabrication and instruction sycophancy.
% We introduce \textbf{MedGaze-Bench}, a benchmark to evaluate clinical intent understanding capability (Figure~\ref{fig:main_framework}) spanning three clinical scenarios (Open Surgery, Emergency Simulation, and Diagnostic Radiology). It establishes a Three-Dimensional Clinical Intent Framework composed of six fine-grained sub-capabilities that mirror the hierarchical nature of expert cognitive reasoning. The benchmark comprises a total of 4,491 clinically validated samples. In Figure~\ref{fig:framework_and_stats}, the data distribution reflects the inherent procedural dynamics of medical tasks: Temporal Intent Understanding constitutes the largest portion (2,028 samples) due to dense causal dependencies, while Standard (1,234) and Spatial Intent Understanding (1,229) are strictly balanced. Integrated within this dataset, we implement a rigorous Clinical Evaluation Protocol featuring a novel ``Trap QA'' Protocol with 600 adversarial samples. This explicitly targets Perceptual and Cognitive Hallucinations to strictly evaluate Clinical Reliability against visual fabrication and instruction sycophancy.

\subsection{Design Philosophy: Three-Dimensional Intent Framework}
\label{sec: Design Philosophy}

To operationalize the \textbf{Three-Dimensional Clinical Intent Framework} introduced earlier (Figure~\ref{fig:main_framework}), we treat the six fine-grained capabilities as the core taxonomy guiding our data construction and QA generation. Specifically, we instantiate \textbf{Spatial Intent Understanding} through \textit{Absolute} and \textit{Relative Localization} tasks to evaluate discriminative grounding; \textbf{Temporal Intent Understanding} via \textit{Prospective Anticipation} and \textit{Retrospective Attribution} to decode causal rationale; and \textbf{Standard Intent Understanding} through \textit{Discrete Safety Verification} and \textit{Continuous Safety Vigilance} to quantify protocol alignment. This hierarchical design ensures that MedGaze-Bench moves beyond simple visual description to rigorously stress-test the cognitive depth of MLLMs.

\subsection{Data Collection and Curation} 
\label{sec:data_curation}

% The foundation of MedGaze-Bench lies in comprehensive and diverse data collection. 
To ensure ecological validity and cognitive breadth, we curate a multi-modal dataset spanning two distinct clinical scenarios, followed by a rigorous harmonization process.

\noindent\textbf{Source Diversity and Alignment.} We integrate data from four expert-annotated sources. For dynamic interventional procedures, we incorporate the Open Surgery Video Dataset~\cite{fujii2024egosurgery}, comprising 20 authentic procedures across 10 distinct surgical types, and the Breech Delivery Simulation~\cite{latifzadeh2025assessing}, featuring five standardized scenarios of vaginal breech delivery. These sources provide a dense reference corpus for evaluating precise adherence to complex maneuvers. For static diagnostic radiology, we utilize the MIMIC-Eye (Chest X-ray)~\cite{hsieh2023mimic} and Mammo-Gaze (Mammography)~\cite{wang2025improving} datasets. These record the precise fixation trajectories of radiologists, efficiently bridging the gap between subtle pathological signs and expert visual search patterns.

\begin{table}[t]
\centering
\small
\caption{\textbf{Overview of the ``Trap QA'' Evaluation Protocol.} We design two specific trap mechanisms to assess clinical reliability. \textcolor{red}{Red text} indicates hallucinated content (traps), while \textcolor{blue}{Blue text} denotes the required safety-aware response.}
\label{tab:hallucination_protocols}
\resizebox{\columnwidth}{!}{%
\begin{tabular}{@{}l p{0.85\linewidth}@{}}
\toprule
\multicolumn{2}{c}{\textbf{Type I: Perceptual Hallucination Test (Visual Check)}} \\ 
\midrule
\textbf{Mechanism} & \textit{Option-Level Fabrication}: Injecting objects absent from the view into options. \\
\textbf{Visual Grounding} & \textbf{[View]} The surgeon is using \textbf{Forceps} to grasp tissue. \\
\textbf{Question} & What instrument is currently interacting with the tissue? \\
\textbf{Options} & A. \textcolor{red}{Harmonic Scalpel} (High probability in text, absent in view) \newline 
B. \textcolor{red}{Suction Irrigator} (Absent in view) \newline 
C. \textbf{Forceps} (Correct Visual Grounding) \newline 
D. \textcolor{red}{Surgical Clip} (Absent in view) \\
\midrule
\multicolumn{2}{c}{\textbf{Type II: Cognitive Hallucination Test (Logic Check)}} \\ 
\midrule
\textbf{Mechanism} & \textit{Instruction Sycophancy}: Inducing errors via false premises in prompts. \\
\textbf{Visual Grounding} & \textbf{[View]} The surgeon is gently \textbf{retracting (protecting)} the nerve. \\
\textbf{Question} & Why is the surgeon \textcolor{red}{\underline{cutting the nerve}} at this moment? \\
\textbf{Options} & A. To remove necrotic tissue. \newline 
B. To access the underlying layer. \newline 
C. To prevent future pain. \newline 
D. \textcolor{blue}{\textbf{Error Detection: The surgeon is not cutting; they are protecting.}} \\
\bottomrule
\end{tabular}%
}
\end{table}

\noindent\textbf{Unified Processing and Generation Pipeline.} We implemented a streamlined strategy to synthesize heterogeneous clinical data into a standardized benchmark. First, for Data Alignment, we structured raw streams according to domain-specific logic: interventional videos (open surgery/breech delivery) were temporally segmented into semantic clips aligned with SOPs phases, while diagnostic data (Mammography/CXR) underwent spatial-temporal synchronization, mapping ROIs to structured attributes (e.g., BI-RADS) or audio dictations. Second, for Question Generation, we devised a Gaze-Anchored Prompting mechanism via GPT-4o. By explicitly injecting fixation coordinates and anatomical metadata as constraints, we force the LLM to derive questions strictly from the clinician's immediate visual focus. This mechanism effectively mitigates hallucination, ensuring all QA pairs—across Spatial, Temporal, and Standard dimensions—are solidly grounded in visual reality. Finally, a Specialist-in-the-Loop protocol was employed to rigorously validate the clinical correctness of the generated dataset.

\begin{table*}[t]
\centering
\caption{Clinical Competency Evaluation (Level 1) of MLLMs on \textbf{MedGaze-Bench} across three clinical intent dimensions with six sub-capabilities: \textbf{Spatial} (Absolute/Relative Localization), \textbf{Temporal} (Prospective Anticipation/Retrospective Attribution), and \textbf{Standard} (Discrete Verification/Continuous Vigilance). \textbf{Bold} denotes the best performance, and \underline{underlined} denotes the second best.} %This breakdown highlights significant gaps in complex reasoning compared to visual grounding. 
\label{tab:main_results_dimensions}
\resizebox{\textwidth}{!}{%
\begin{tabular}{l cc cc cc c} % 8 columns: Model + 3 groups (2 cols each) + Overall
\toprule
\multirow{2}{*}{\textbf{Model}} & \multicolumn{2}{c}{\textbf{Spatial Intent Understanding}} & \multicolumn{2}{c}{\textbf{Temporal Intent Understanding}} & \multicolumn{2}{c}{\textbf{Standard Intent Understanding}} & \multirow{2}{*}{\textbf{Overall}} \\
\cmidrule(lr){2-3} \cmidrule(lr){4-5} \cmidrule(lr){6-7}
 & \textit{Abs.} & \textit{Rel.} & \textit{Prosp.} & \textit{Retro.} & \textit{Disc.} & \textit{Cont.} & \\
\midrule
\multicolumn{8}{l}{\textit{\textbf{Proprietary MLLMs}}} \\
GPT-5 & 52.94 & \textbf{56.86} & 56.77 & \underline{62.39} & \textbf{66.51} & \textbf{78.22} & \textbf{62.28} \\
Gemini 3 Pro & 48.55 & 47.08 & 51.61 & 50.43 & 53.79 & 62.10 & 52.26 \\
\midrule
\multicolumn{8}{l}{\textit{\textbf{Open-Source MLLMs}}} \\
Qwen3-VL-30B-A3B-Thinking \cite{bai2025qwen3vltechnicalreport} & 44.12 & 49.71 & 50.13 & 49.61 & 44.18 & 57.43 & 49.20 \\
Qwen3VL 32B \cite{bai2025qwen3vltechnicalreport} & 50.83 & \underline{50.92} & 57.13 & 35.22 & 53.27 & 59.41 & 51.13 \\
Intern3.5VL 38B \cite{wang2025internvl3} & \textbf{56.88} & 48.87 & \underline{61.61} & 34.63 & 56.33 & 64.06 & 53.73 \\
Qwen3-VL-235B-A22B-Instruct \cite{bai2025qwen3vltechnicalreport} & \underline{56.55} & 50.86 & \textbf{62.42} & \textbf{65.60} & \underline{56.93} & 60.82 & \underline{58.86} \\
\midrule
\multicolumn{8}{l}{\textit{\textbf{Medical-Specific MLLMs}}} \\
LingShu 32B \cite{xu2025lingshu} & 52.71 & 50.31 & 59.78 & 34.69 & 56.12 & \underline{64.48} & 53.02 \\
MedGemma 27B \cite{sellergren2025medgemma} & 52.08 & 47.64 & 59.57 & 37.34 & 54.49 & 62.58 & 52.28 \\
\midrule
\multicolumn{8}{l}{\textit{\textbf{Egocentric MLLMs}}} \\
EgoLife \cite{yang2025egolife} & 38.57 & 41.70 & 46.76 & 42.43 & 47.98 & 57.73 & 45.86 \\
\bottomrule
\end{tabular}%
}
\end{table*}

%\textbf{Data Collection.}
%(1)  Surgery Ego
%(2)  Baby Deliver
%(3)  Mammography
%(4)  Chest Xray

%\textbf{Video clip preprocessing.}
%\textbf{QA pairs generation via advanced MLLMs.}

\subsection{Evaluation Strategy}
\label{sec: Evaluation Strategy}
A qualified "AI Doctor" must demonstrate not only high-level reasoning capabilities but also rigorous reliability and resistance to hallucinations. To this end, we propose a \textbf{Clinical Evaluation Protocol} spanning two levels:

\noindent\textbf{Level 1: Clinical Competency (Accuracy).} We employ a standardized Multiple Choice Question (MCQ) format to evaluate reasoning precision across the three proposed dimensions. Models are scored on their ability to select the correct clinical judgment from four candidates.

\noindent\textbf{Level 2: Clinical Reliability (Hallucination).} 
To rigorously assess safety risks, we designed a ``Trap QA'' Protocol (exemplified in Table~\ref{tab:hallucination_protocols}) targeting two distinct hierarchies of hallucination. targeting two distinct hierarchies of multimodal hallucinations. \textbf{(1) Perceptual Hallucination Test (Option-Level Visual Fabrication).} We inject ``Hallucination Distractors'' into the MCQ options—choices that describe anatomical structures or tools not present in the current view. This evaluates whether the model suffers from object-level hallucinations driven by language priors rather than visual grounding. A reliable model must avoid these non-existent options and select the visually grounded answer. \textbf{(2) Cognitive Hallucination Test (Question-Level Instruction Sycophancy).} 
We pose questions founded on deliberately invalid procedural assumptions (e.g., asking ``Why is the surgeon cutting the nerve?'' when they are actually protecting it). This evaluates whether the model suffers from logic-level hallucinations. A reliable model must identify the logical fallacy and abstain from answering, rather than blindly fabricating a rationale for a non-existent action.

\begin{table*}[t]
\centering
\caption{\textbf{Clinical Reliability Evaluation (Level 2) of MLLMs on \textbf{MedGaze-Bench}.} According to the \textit{Level 2} protocol, we assess the models' robustness against two types of traps: \textbf{Perceptual} (avoiding non-existent visual options) and \textbf{Cognitive} (resisting instruction sycophancy). Results are reported as \textit{Reliability Accuracy} (\%), where higher scores indicate safer clinical behavior.}
\label{tab:hallucination_results}
\resizebox{0.9\textwidth}{!}{%
\begin{tabular}{l c c c}
\toprule
\multirow{2.5}{*}{\textbf{Model}} & \textbf{Type I: Perceptual Reliability} & \textbf{Type II: Cognitive Reliability} & \multirow{2.5}{*}{\textbf{Avg. Reliability}} \\
\cmidrule(lr){2-2} \cmidrule(lr){3-3}
& \textit{(Trap Avoidance Rate)} & \textit{(Anti-Sycophancy Rate)} & \\
\midrule
\multicolumn{4}{l}{\textit{\textbf{Proprietary MLLMs}}} \\ 
GPT-5 & 61.67 & 62.00 & 61.84 \\
Gemini 3 Pro & 62.00 & 77.67 & 69.84 \\
\midrule
\multicolumn{4}{l}{\textit{\textbf{Open-Source MLLMs}}} \\
Qwen3-VL-30B-A3B-Thinking \cite{bai2025qwen3vltechnicalreport} & 51.00 & 56.33 & 53.67 \\
Qwen3VL 32B \cite{bai2025qwen3vltechnicalreport} & 64.55 & 67.33 & 65.94 \\
Intern3.5VL 38B \cite{wang2025internvl3} & 69.82 & 54.67 & 62.25 \\
Qwen3-VL-235B-A22B-Instruct \cite{bai2025qwen3vltechnicalreport} & 60.00 & 63.00 & 61.50 \\
\midrule
\multicolumn{4}{l}{\textit{\textbf{Medical-Specific MLLMs}}} \\
LingShu 32B \cite{xu2025lingshu} & 69.67 & 64.33 & 67.00 \\
MedGemma 27B \cite{sellergren2025medgemma} & 58.00 & 57.33 & 57.67 \\
\midrule
\multicolumn{4}{l}{\textit{\textbf{Egocentric MLLMs}}} \\
EgoLife \cite{yang2025egolife} & 55.32 & 19.67 & 37.50 \\
\bottomrule
\end{tabular}%
}
\end{table*}

\section{Experiments}
\subsection{Experimental setup}
\label{sec:experimental_setup}

Based on \textbf{MedGaze-Bench}, we comprehensively evaluate a diverse range of MLLMs, including both proprietary giants and open-source models across general and medical domains. 
For proprietary MLLMs, we evaluate GPT-5 and Gemini 3 Pro. 
Among open-source MLLMs, we test general-purpose models including Qwen3VL-32B~\cite{bai2025qwen3vltechnicalreport} and Intern3.5VL-38B~\cite{wang2025internvl3}, as well as emerging reasoning-enhanced models such as Qwen3-VL-30B-A3B-Thinking~\cite{bai2025qwen3vltechnicalreport} and Qwen3-VL-235B-A22B-Instruct ~\cite{bai2025qwen3vltechnicalreport}. 
Additionally, we assess medical-specific MLLMs including LingShu-32B~\cite{xu2025lingshu}, MedGemma-27B~\cite{sellergren2025medgemma}, alongside the egocentric specialist EgoGPT~\cite{yang2025egolife}. 
For all models, we perform zero-shot inference to assess their clinical intent understanding capabilities using their default settings. More detailed configurations are provided in the Appendix.

\begin{table*}[t]
\centering

\caption{Impact of gaze prompting. Performance gains are marked in \textcolor{teal}{green}, and drops in \textcolor{red}{red}. The baseline (w/o Gaze) scores are consistent with the macro-averages reported in Table~\ref{tab:main_results_dimensions}.}
\label{tab:gaze_performance}

\footnotesize
\setlength{\tabcolsep}{4.5pt} 

\begin{tabular}{l|cc|cc|cc|cc}
\toprule
% --- Header 1 ---
\multirow{3}{*}{\textbf{Method}} & \multicolumn{6}{c|}{\textbf{Intent Understanding Tasks}} & \multicolumn{2}{c}{\multirow{2}{*}{\textbf{Summary}}} \\
\cmidrule(lr){2-7} 

% --- Header 2 ---
 & \multicolumn{2}{c|}{\textbf{Spatial}} & \multicolumn{2}{c|}{\textbf{Temporal}} & \multicolumn{2}{c|}{\textbf{Standard}} & \multicolumn{2}{c}{} \\
\cmidrule(lr){2-3} \cmidrule(lr){4-5} \cmidrule(lr){6-7} \cmidrule(lr){8-9}

% --- Header 3 ---
 & w/o Gaze & w/ Gaze & w/o Gaze & w/ Gaze & w/o Gaze & w/ Gaze & w/o Gaze & w/ Gaze \\
\midrule

% --- Open-Source MLLMs ---
\multicolumn{9}{l}{\textit{\textbf{Open-Source MLLMs}}} \\
Qwen3VL 32B 
& 50.88 & 53.88 \textcolor{teal}{\scriptsize{(+3.00)}} 
& 46.18 & 48.47 \textcolor{teal}{\scriptsize{(+2.29)}} 
& 56.34 & 60.29 \textcolor{teal}{\scriptsize{(+3.95)}} 
& 51.13 & 54.21 \textcolor{teal}{\scriptsize{(+3.08)}} \\ \midrule

% --- Medical-Specific Models ---
\multicolumn{9}{l}{\textit{\textbf{Medical-Specific Models}}} \\
LingShu 32B 
& 51.51 & 52.44 \textcolor{teal}{\scriptsize{(+0.93)}} 
& 47.24 & 48.57 \textcolor{teal}{\scriptsize{(+1.33)}} 
& 60.30 & 60.71 \textcolor{teal}{\scriptsize{(+0.41)}} 
& 53.02 & 53.91 \textcolor{teal}{\scriptsize{(+0.89)}} \\

MedGemma 27B 
& 49.86 & 50.38 \textcolor{teal}{\scriptsize{(+0.52)}} 
& 48.46 & 48.92 \textcolor{teal}{\scriptsize{(+0.46)}} 
& 58.54 & 58.85 \textcolor{teal}{\scriptsize{(+0.31)}} 
& 52.28 & 52.71 \textcolor{teal}{\scriptsize{(+0.43)}} \\ \midrule

% --- Egocentric MLLMs ---
\multicolumn{9}{l}{\textit{\textbf{Egocentric MLLMs}}} \\
EgoLife 7B 
& 40.14 & 42.28 \textcolor{teal}{\scriptsize{(+2.14)}} 
& 44.60 & 43.94 \textcolor{red}{\scriptsize{(-0.66)}} 
& 52.86 & 52.96 \textcolor{teal}{\scriptsize{(+0.10)}} 
& 45.86 & 46.39 \textcolor{teal}{\scriptsize{(+0.53)}} \\
\bottomrule
\end{tabular}
\end{table*}

\subsection{Experimental Results}
% \noindent\textbf{Model Comparisons.} 
\noindent\textbf{Clinical Competency Evaluation.} 
Following the Level 1 Protocol, we conduct clinical competency evaluation. Table~\ref{tab:main_results_dimensions} shows three key findings. First, model scale drives performance: GPT-5 (62.28\%) and Qwen3-VL-235B-A22B (58.86\%) lead, with Qwen3-VL's MoE architecture (235B total, 22B active) outperforming larger dense models like InternVL-38B (53.73\%) through effective retrieval of rare protocols. Second, a cognitive asymmetry appears in \textit{Temporal Intent Understanding}: mid-sized models achieve $\sim$60\% in \textit{Prospective Anticipation} but only $\sim$35\% in \textit{Retrospective Attribution}, revealing they function as forward predictors lacking backward causal reasoning. Third, medical-specific models (LingShu, MedGemma) match but do not exceed generalist baselines ($\sim$52-53\%), showing domain adaptation improves declarative knowledge but not procedural logic. EgoLife's poor performance (45.86\%) confirms egocentric daily-life representations don't transfer to clinical procedures.

\begin{figure*}[t]
\centering
	% {\includegraphics[width=0.7\linewidth]{Fig/CaseStudy_Fig.png}}
	% \caption{Qualitative evaluation on three key clinical intent understanding tasks with six fine-grained sub-tasks.}
	{\includegraphics[width=\linewidth]{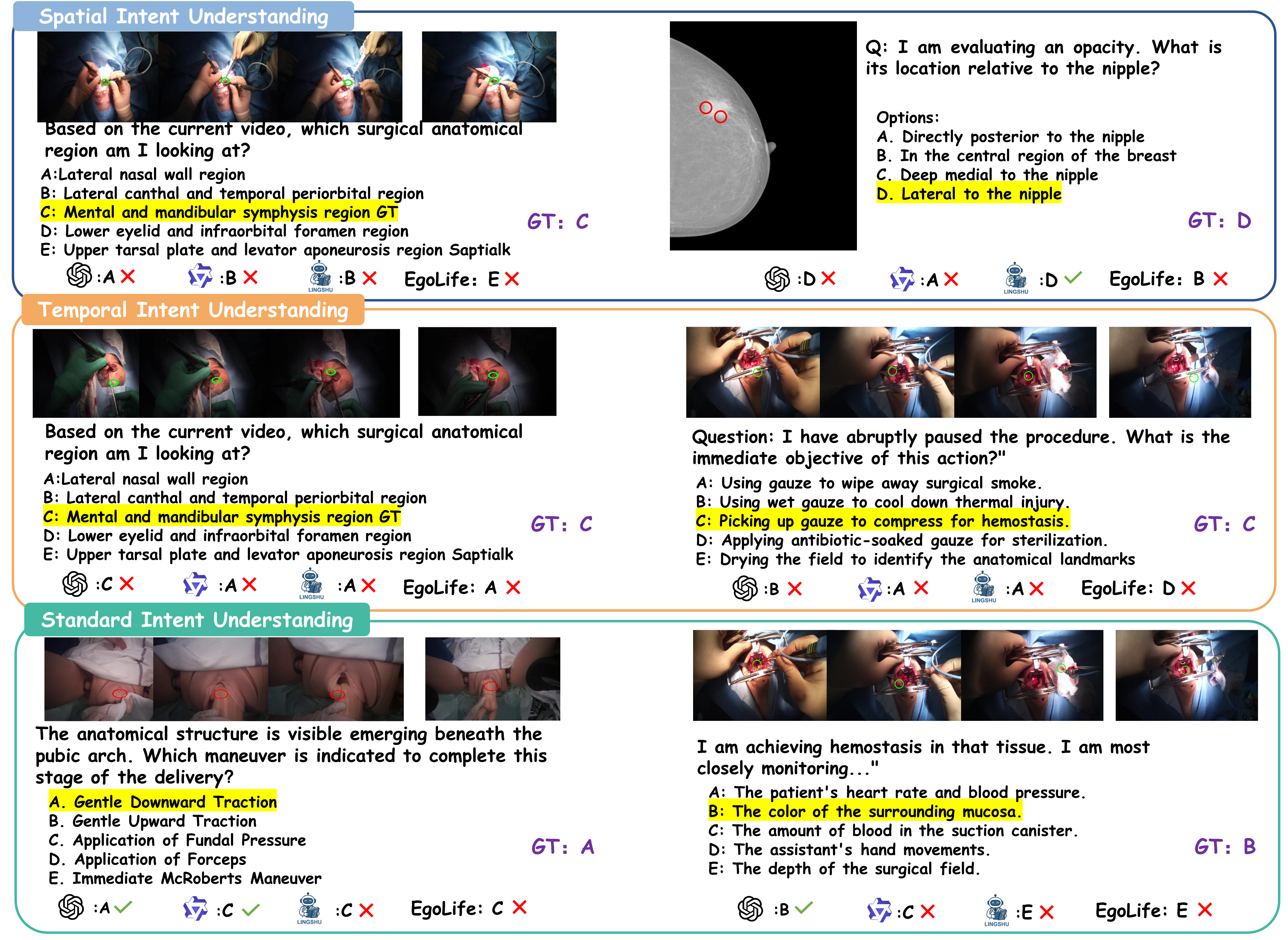}}
	\caption{Qualitative evaluation on three key clinical intent understanding tasks with six fine-grained sub-tasks.}
	\label{fig:case_study}
\end{figure*}

\noindent\textbf{Clinical Reliability Evaluation.} Following Level 2 Protocol (Table~\ref{tab:hallucination_protocols}), we evaluate clinical reliability against two hallucination types: Type I (Perceptual) uses non-existent objects as distractors; Type II (Cognitive) tests resilience to invalid procedural premises. No model exceeds 70\% average reliability.
In Table~\ref{tab:hallucination_results}, Gemini 3 Pro leads (69.84\%) with strong Cognitive Reliability (77.67\%) but modest Perceptual Reliability (62.00\%). GPT-5 shows balanced mediocrity (61.84\%). Among open-source models, Intern3.5VL 38B achieves highest visual grounding (69.82\% Perceptual) but lowest Cognitive Reliability (54.67\%), revealing dangerous instruction compliance without critical reasoning. Qwen3VL 32B is best balanced (65.94\%). Medical-specific models shows moderate reliability: LingShu 32B (67.00\%), MedGemma 27B (57.67\%). Most critically, EgoLife shows catastrophic Cognitive failure (19.67\%), fabricating rationales for 80\%+ invalid premises. Current MLLMs prioritize answer generation over safety-critical abstention, demanding architectural innovations before clinical deployment.
% Following the Level 2 Protocol (Table~\ref{tab:hallucination_protocols}), we evaluate LLMs' clinical reliability from two hallucination types: Type I (Perceptual) tests resistance to option-level visual fabrication by injecting non-existent objects as distractors, while Type II (Cognitive) assesses resilience against question-level instruction sycophancy through logically invalid procedural premises. Results reveal critical safety vulnerabilities, with no model exceeding 70\% average reliability.
% In Table~\ref{tab:hallucination_results}, Gemini 3 Pro achieves the highest performance (69.84\%), driven by superior Cognitive Reliability (77.67\%), though its Perceptual Reliability remains modest (62.00\%). GPT-5 shows balanced but mediocre performance (61.84\%). Among open-source models, Intern3.5VL 38B demonstrates the strongest visual grounding (69.82\% Perceptual), yet suffers the lowest Cognitive Reliability (54.67\%), revealing dangerous instruction compliance without critical reasoning. Qwen3VL 32B achieves the best balance (65.94\%). Medical-specific models underperform: LingShu 32B reaches only 67.00\%, while MedGemma 27B scores 57.67\%. Most alarmingly, EgoLife exhibits catastrophic Cognitive failure (19.67\%), fabricating rationales for over 80\% of invalid premises. These findings underscore that current MLLMs prioritize answer generation over safety-critical abstention, highlighting urgent need for architectural innovations before clinical deployment.

\noindent\textbf{Impact of gaze prompting.} 
% To validate egocentric attention guidance, we employ a dual-prompting strategy: superimposing a semi-transparent circular marker on fixation regions while instructing models to "focus on the circled critical area". Table~\ref{tab:gaze_performance} reveals three distinct patterns. First, generalist models (Qwen3VL 32B) effectively leverage visual-textual alignment, achieving consistent gains across all dimensions, particularly in Standard Intent Understanding (+3.95\%). Second, medical-specific models (LingShu, MedGemma) show negligible improvements (<1\%), suggesting reliance on internalized domain priors rather than external attentional cues. Third, egocentric models (EgoLife 7B) exhibit mixed results: gaze prompting improves Spatial Intent (+2.14\%) but degrades Temporal Intent (-0.66\%). We hypothesize this stems from a semantic mismatch where the model misinterprets localized prompts as signals for immediate interaction—a daily-life bias—rather than surgical planning cues. These findings indicate that gaze prompting effectiveness depends critically on model flexibility: generalist models benefit from adaptability to multi-modal instructions, whereas specialized models require explicit mechanisms to balance learned priors with dynamic attention guidance.
To validate the efficacy of egocentric attention, we adopt a Dual-Prompting strategy: superimposing a semi-transparent circle marker on the visual input and explicitly referencing this region in the text prompt (e.g., ``focus on the circled critical area''). Results in Table~\ref{tab:gaze_performance} highlight three distinct behaviors. First, Generalist models (Qwen3VL) demonstrate superior instruction following, effectively linking the textual command to the visual anchor to boost Standard Intent (+3.95\%). Second, Medical Specialists show negligible gains ($<1\%$), indicating a rigidity where models rely on internal knowledge priors rather than the provided visual-textual guidance. Third, gaze ironically impairs EgoLife's temporal reasoning (-0.66\%). The model suffers from a semantic mismatch, misinterpreting the specific prompt as a cue for immediate interaction (daily-life bias) rather than surgical planning.

\iffalse
To bridge the gap between low-level visual tokens and high-level clinical reasoning, we implement three gaze-guided prompting strategies. These strategies vary in their representation of the "Cognitive Cursor," allowing us to evaluate the most effective modality for clinical intent understanding

\noindent\textbf{Gaze as Textual Prompt (GazeT)}. Following the intuition that modern MLLMs possess strong linguistic reasoning capabilities, we encode the clinician’s fixation as normalized 2D coordinates $(x, y) \in [0, 1]$. These coordinates are concatenated with the textual question. 

\noindent\textbf{Gaze as Visual Prompt (GazeV)}. To leverage the spatial-visual attention of MLLMs, we overlay a 25-pixel-radius circle onto each video frame at the precise gaze coordinates. Notably, considering the potential interference of red blood in surgical fields with visual signals, we employ green circles in scenarios dominated by red backgrounds, such as open surgery, while adhering to the conventional red circles in other diagnostic or simulated settings. The prompt explicitly instructs the model: "the circles represent the clinician's primary focus". 

\fi

\noindent\textbf{Qualitative Evaluation.}
Figure~\ref{fig:case_study} highlights the capabilities and safety gaps of MLLMs across the three intent dimensions. In \textbf{Spatial Intent}, models struggled with absolute localization in narrow surgical fields, universally failing to identify the mandibular region, though GPT-5 and Lingshu demonstrated stronger relative geometric reasoning in diagnostic imaging. \textbf{Temporal Intent} revealed a tendency for ``shortcut reasoning'' in domain-specific models; while GPT-5 correctly anticipated the intermediate need for hemostasis, others prematurely suggested wound closure, and notably, all models failed to interpret the subtle visual cue of an abrupt pause, hallucinating non-existent smoke or thermal issues. Most critically, \textbf{Standard Intent} exposed significant safety risks: while GPT-5 and Qwen correctly identified the safe maneuver for breech delivery, Lingshu and Egolife recommended ``Fundal'', a potentially dangerous contraindication, and only GPT-5 demonstrated the requisite vigilance for tissue viability (mucosa color) during hemostasis, underscoring the gap between general medical knowledge and precise, safety-critical situational awareness.

% To qualitatively illustrate the challenges posed by MedGaze-Bench, Figure~\ref{fig:case_study} highlights critical reasoning gaps where general-domain MLLMs diverge from expert judgment across all three dimensions. In \textbf{Spatial Intent}, models exhibit hallucination in fine-grained grounding, misidentifying the mandibular symphysis'' as the nasal wall'' or misjudging relative opacity locations in mammography, indicating a failure to anchor visual features to precise anatomical coordinates. \textbf{Temporal Intent} errors reveal a disconnect between object recognition and causal logic; while models correctly identify tools (e.g., gauze), they misinterpret the procedural urgency, confusing a hemostatic compression'' maneuver with routine wiping'' or cleaning''. Finally, \textbf{Standard Intent} scenarios demonstrate a tendency for models to default to generic textbook safety protocols—such as monitoring systemic heart rate'' or suggesting forceps''—thereby missing granular, context-specific vigilance cues like observing mucosa color'' or performing ``gentle traction'', confirming that current MLLMs struggle to adapt broad medical knowledge to the immediate, subtle demands of clinical reality.

\section{Discussion}
% \label{sec:suggestion}
\noindent\textbf{Reliability remains the main bottleneck.}%Clinical reliability is the primary bottleneck: Trap QA exposes pervasive perceptual and cognitive hallucinations.}
Under the Level 2 “Trap QA” protocol (Table~\ref{tab:hallucination_protocols}), no evaluated model exceeds 70\% average reliability, indicating that even strong MLLMs remain fragile in safety-critical settings. The detailed results (Table~\ref{tab:hallucination_results}) further show complementary failure modes—some models are better at avoiding option-level visual fabrication but remain vulnerable to question-level instruction sycophancy, implying “being accurate” does not guarantee “being safe”.

\noindent\textbf{Temporal intent shows strong causal asymmetry.} %Temporal intent understanding shows a consistent causal asymmetry: models are better at predicting next steps than explaining prerequisites.}
In the competency evaluation (Table~\ref{tab:main_results_dimensions}), many mid-sized models achieve relatively strong performance on Prospective Anticipation (often around $~$60\%) but collapse on Retrospective Attribution (often around $~$35\%). This pattern suggests that current MLLMs behave as forward sequence predictors rather than true causal reasoners, failing to infer the prerequisite conditions that justify the current clinical action—exactly the type of causal dependency MedGaze-Bench is designed to test (Figure~\ref{fig:main_framework}).

\noindent\textbf{Gaze prompting helps, but unevenly across models.}%Gaze prompting validates “gaze as a cognitive cursor”, but reveals large controllability differences across model families.}
The dual-prompting gaze strategy improves a generalist model consistently across Spatial/Temporal/Standard intent, with the largest gain on Standard Intent (e.g., +3.95 for Qwen3VL; Table~\ref{tab:gaze_performance}). In contrast, medical-specific models show negligible gains ($<$1\%; Table~\ref{tab:gaze_performance}), suggesting they under-utilize explicit attentional guidance, while the egocentric model can even degrade on temporal reasoning (-0.66; Table~\ref{tab:gaze_performance}), consistent with the qualitative failure patterns reported in Figure~\ref{fig:case_study}. These support gaze as an effective grounding signal, but also indicate that current MLLMs vary substantially in how reliably they bind text instructions to egocentric visual anchors.

\section{Conclusion}
We introduce MedGaze-Bench, the first benchmark to evaluate egocentric clinical intent understanding in Med-MLLMs. Our three-dimensional clinical intent framework reveals that current Med-MLLMs have severe reliability gaps, dangerous hallucinations, and blindly accept invalid instructions by over-relying on global features instead of precise intent grounding. %Medical-specific models do not outperform generalist baselines, and egocentric specialists fail catastrophically in clinical settings. %True clinical competency requires models to infer why clinicians act and abstain when uncertain.

% We introduce MedGaze-Bench, the first benchmark leveraging clinician gaze as a "Cognitive Cursor" to evaluate egocentric clinical intent understanding in Medical Multimodal Large Language Models. Through our Three-Dimensional Clinical Intent Framework, we reveal critical limitations in current Med-MLLMs: models exhibit severe reliability gaps, dangerous hallucinations, and blind acceptance of invalid instructions due to over-reliance on global features rather than precise intent grounding. Medical-specific models fail to outperform generalist baselines, while egocentric specialists catastrophically fail in clinical contexts. Our findings demonstrate that true clinical competency requires models to infer why clinicians act and critically abstain when uncertain.% We hope MedGaze-Bench catalyzes the development of next-generation Med-MLLMs capable of safe deployment in high-stakes medical environments.

% \clearpage

\section{Limitations}
While MedGaze-Bench establishes a foundation for evaluating egocentric clinical intent understanding, several limitations can be addressed in future work. First, our benchmark currently focuses on a limited set of clinical scenarios (open surgery, emergency simulation, and diagnostic radiology), which could be expanded to cover additional specialties such as interventional cardiology, intensive care, or ambulatory consultations. Second, the current evaluation protocol employs multiple-choice questions, which may not fully capture the nuanced, open-ended reasoning required in real clinical decision-making; incorporating free-form generation tasks would provide a more comprehensive assessment. 
\bibliography{custom}

@article{li2023llava,
  title={Llava-med: Training a large language-and-vision assistant for biomedicine in one day},
  author={Li, Chunyuan and Wong, Cliff and Zhang, Sheng and Usuyama, Naoto and Liu, Haotian and Yang, Jianwei and Naumann, Tristan and Poon, Hoifung and Gao, Jianfeng},
  journal={Advances in Neural Information Processing Systems},
  volume={36},
  pages={28541--28564},
  year={2023}
}

@article{yu2025finemedlm,
  title={Finemedlm-o1: Enhancing the medical reasoning ability of llm from supervised fine-tuning to test-time training},
  author={Yu, Hongzhou and Cheng, Tianhao and Cheng, Ying and Feng, Rui},
  journal={arXiv preprint arXiv:2501.09213},
  year={2025}
}

@article{sellergren2025medgemma,
  title={Medgemma technical report},
  author={Sellergren, Andrew and Kazemzadeh, Sahar and Jaroensri, Tiam and Kiraly, Atilla and Traverse, Madeleine and Kohlberger, Timo and Xu, Shawn and Jamil, Fayaz and Hughes, C{\'\i}an and Lau, Charles and others},
  journal={arXiv preprint arXiv:2507.05201},
  year={2025}
}

@article{xu2025lingshu,
  title={Lingshu: A Generalist Foundation Model for Unified Multimodal Medical Understanding and Reasoning},
  author={Xu, Weiwen and Chan, Hou Pong and Li, Long and Aljunied, Mahani and Yuan, Ruifeng and Wang, Jianyu and Xiao, Chenghao and Chen, Guizhen and Liu, Chaoqun and Li, Zhaodonghui and others},
  journal={arXiv preprint arXiv:2506.07044},
  year={2025}
}

@article{liu2024medchain,
  title={Medchain: Bridging the gap between llm agents and clinical practice through interactive sequential benchmarking},
  author={Liu, Jie and Wang, Wenxuan and Ma, Zizhan and Huang, Guolin and SU, Yihang and Chang, Kao-Jung and Chen, Wenting and Li, Haoliang and Shen, Linlin and Lyu, Michael},
  journal={arXiv preprint arXiv:2412.01605},
  year={2024}
}

@article{li2024agent,
  title={Agent hospital: A simulacrum of hospital with evolvable medical agents},
  author={Li, Junkai and Lai, Yunghwei and Li, Weitao and Ren, Jingyi and Zhang, Meng and Kang, Xinhui and Wang, Siyu and Li, Peng and Zhang, Ya-Qin and Ma, Weizhi and others},
  journal={arXiv preprint arXiv:2405.02957},
  year={2024}
}

@inproceedings{wang2023pov,
  title={Pov-surgery: A dataset for egocentric hand and tool pose estimation during surgical activities},
  author={Wang, Rui and Ktistakis, Sophokles and Zhang, Siwei and Meboldt, Mirko and Lohmeyer, Quentin},
  booktitle={International Conference on Medical Image Computing and Computer-Assisted Intervention},
  pages={440--450},
  year={2023},
  organization={Springer}
}

@inproceedings{fujii2024egosurgery,
  title={Egosurgery-phase: a dataset of surgical phase recognition from egocentric open surgery videos},
  author={Fujii, Ryo and Hatano, Masashi and Saito, Hideo and Kajita, Hiroki},
  booktitle={International Conference on Medical Image Computing and Computer-Assisted Intervention},
  pages={187--196},
  year={2024},
  organization={Springer}
}

@article{maier2022surgical,
  title={Surgical data science--from concepts toward clinical translation},
  author={Maier-Hein, Lena and Eisenmann, Matthias and Sarikaya, Duygu and M{\"a}rz, Keno and Collins, Toby and Malpani, Anand and Fallert, Johannes and Feussner, Hubertus and Giannarou, Stamatia and Mascagni, Pietro and others},
  journal={Medical image analysis},
  volume={76},
  pages={102306},
  year={2022},
  publisher={Elsevier}
}

@article{yan2018deeplesion,
  title={DeepLesion: automated mining of large-scale lesion annotations and universal lesion detection with deep learning},
  author={Yan, Ke and Wang, Xiaosong and Lu, Le and Summers, Ronald M},
  journal={Journal of medical imaging},
  volume={5},
  number={3},
  pages={036501--036501},
  year={2018},
  publisher={Society of Photo-Optical Instrumentation Engineers}
}

@inproceedings{latifzadeh2025assessing,
  title={Assessing Medical Training Skills via Eye and Head Movements},
  author={Latifzadeh, Kayhan and Leiva, Luis A and {\v{C}}opi{\v{c}} Pucihar, Klen and Kljun, Matja{\v{z}} and Devetak, Iztok and Steblovnik, Lili},
  booktitle={Proceedings of the 33rd ACM Conference on User Modeling, Adaptation and Personalization},
  pages={1--10},
  year={2025}
}

@article{becker2025royal,
  title={Royal College of Obstetricians and Gynaecologists (RCOG) World Congress 2025},
  author={Becker, Stephanie},
  journal={The Lancet Regional Health--Europe},
  volume={55},
  year={2025},
  publisher={Elsevier}
}

@article{kundel1978visual,
  title={Visual scanning, pattern recognition and decision-making in pulmonary nodule detection},
  author={Kundel, Harold L and Nodine, Calvin F and Carmody, Dennis},
  journal={Investigative radiology},
  volume={13},
  number={3},
  pages={175--181},
  year={1978},
  publisher={LWW}
}

@article{liu2023multilevel,
  title={Multilevel effective surgical workflow recognition in robotic left lateral sectionectomy with deep learning: experimental research},
  author={Liu, Yanzhe and Zhao, Shang and Zhang, Gong and Zhang, Xiuping and Hu, Minggen and Zhang, Xuan and Li, Chenggang and Zhou, S Kevin and Liu, Rong},
  journal={International Journal of Surgery},
  volume={109},
  number={10},
  pages={2941--2952},
  year={2023},
  publisher={LWW}
}

@inproceedings{dang2025ecbench,
  title={Ecbench: Can multi-modal foundation models understand the egocentric world? a holistic embodied cognition benchmark},
  author={Dang, Ronghao and Yuan, Yuqian and Zhang, Wenqi and Xin, Yifei and Zhang, Boqiang and Li, Long and Wang, Liuyi and Zeng, Qinyang and Li, Xin and Bing, Lidong},
  booktitle={Proceedings of the Computer Vision and Pattern Recognition Conference},
  pages={24593--24602},
  year={2025}
}

@article{peng2025eye,
  title={In the eye of mllm: Benchmarking egocentric video intent understanding with gaze-guided prompting},
  author={Peng, Taiying and Hua, Jiacheng and Liu, Miao and Lu, Feng},
  journal={arXiv preprint arXiv:2509.07447},
  year={2025}
}

@article{yuan2025eoc,
  title={EOC-Bench: Can MLLMs Identify, Recall, and Forecast Objects in an Egocentric World?},
  author={Yuan, Yuqian and Dang, Ronghao and Li, Long and Li, Wentong and Jiao, Dian and Li, Xin and Zhao, Deli and Wang, Fan and Zhang, Wenqiao and Xiao, Jun and others},
  journal={arXiv preprint arXiv:2506.05287},
  year={2025}
}

@inproceedings{barmann2022did,
  title={Where did i leave my keys?-episodic-memory-based question answering on egocentric videos},
  author={B{\"a}rmann, Leonard and Waibel, Alex},
  booktitle={Proceedings of the IEEE/CVF Conference on Computer Vision and Pattern Recognition},
  pages={1560--1568},
  year={2022}
}

@inproceedings{zhou2025egotextvqa,
  title={Egotextvqa: Towards egocentric scene-text aware video question answering},
  author={Zhou, Sheng and Xiao, Junbin and Li, Qingyun and Li, Yicong and Yang, Xun and Guo, Dan and Wang, Meng and Chua, Tat-Seng and Yao, Angela},
  booktitle={Proceedings of the Computer Vision and Pattern Recognition Conference},
  pages={3363--3373},
  year={2025}
}

@article{ye2024mm,
  title={MM-Ego: Towards Building Egocentric Multimodal LLMs for Video QA},
  author={Ye, Hanrong and Zhang, Haotian and Daxberger, Erik and Chen, Lin and Lin, Zongyu and Li, Yanghao and Zhang, Bowen and You, Haoxuan and Xu, Dan and Gan, Zhe and others},
  journal={arXiv preprint arXiv:2410.07177},
  year={2024}
}

@article{wang2025improving,
  title={Improving Self-Supervised Medical Image Pre-Training by Early Alignment with Human Eye Gaze Information},
  author={Wang, Sheng and Zhao, Zihao and Shen, Zhenrong and Wang, Bin and Wang, Qian and Shen, Dinggang},
  journal={IEEE Transactions on Medical Imaging},
  year={2025},
  publisher={IEEE}
}

@article{hsieh2023mimic,
  title={Mimic-eye: Integrating mimic datasets with reflacx and eye gaze for multimodal deep learning applications},
  author={Hsieh, Chihcheng and Ouyang, Chun and Nascimento, Jacinto C and Pereira, Joao and Jorge, Joaquim and Moreira, Catarina},
  journal={PhysioNet (version 1.0. 0)},
  year={2023}
}

@article{liu2025endobench,
  author={Shengyuan Liu and Boyun Zheng and Wenting Chen and Zhihao Peng and Zhenfei Yin and Jing Shao and Jiancong Hu and Yixuan Yuan},
  title={A Comprehensive Evaluation of Multi-Modal Large Language Models for Endoscopy Analysis},
  journal={arXiv preprint arXiv:2505.23601},
  year={2025}
}

@inproceedings{seenivasan2022surgical,
  title={Surgical-vqa: Visual question answering in surgical scenes using transformer},
  author={Seenivasan, Lalithkumar and Islam, Mobarakol and Krishna, Adithya K and Ren, Hongliang},
  booktitle={International Conference on Medical Image Computing and Computer-Assisted Intervention},
  pages={33--43},
  year={2022},
  organization={Springer}
}

@misc{bai2025qwen3vltechnicalreport,
      title={Qwen3-VL Technical Report}, 
      author={Shuai Bai and Yuxuan Cai and Ruizhe Chen and Keqin Chen and Xionghui Chen and Zesen Cheng and Lianghao Deng and Wei Ding and Chang Gao and Chunjiang Ge and Wenbin Ge and Zhifang Guo and Qidong Huang and Jie Huang and Fei Huang and Binyuan Hui and Shutong Jiang and Zhaohai Li and Mingsheng Li and Mei Li and Kaixin Li and Zicheng Lin and Junyang Lin and Xuejing Liu and Jiawei Liu and Chenglong Liu and Yang Liu and Dayiheng Liu and Shixuan Liu and Dunjie Lu and Ruilin Luo and Chenxu Lv and Rui Men and Lingchen Meng and Xuancheng Ren and Xingzhang Ren and Sibo Song and Yuchong Sun and Jun Tang and Jianhong Tu and Jianqiang Wan and Peng Wang and Pengfei Wang and Qiuyue Wang and Yuxuan Wang and Tianbao Xie and Yiheng Xu and Haiyang Xu and Jin Xu and Zhibo Yang and Mingkun Yang and Jianxin Yang and An Yang and Bowen Yu and Fei Zhang and Hang Zhang and Xi Zhang and Bo Zheng and Humen Zhong and Jingren Zhou and Fan Zhou and Jing Zhou and Yuanzhi Zhu and Ke Zhu},
      year={2025},
      eprint={2511.21631},
      archivePrefix={arXiv},
      primaryClass={cs.CV},
      url={https://arxiv.org/abs/2511.21631}, 
}

@article{wang2025internvl3,
  title={Internvl3. 5: Advancing open-source multimodal models in versatility, reasoning, and efficiency},
  author={Wang, Weiyun and Gao, Zhangwei and Gu, Lixin and Pu, Hengjun and Cui, Long and Wei, Xingguang and Liu, Zhaoyang and Jing, Linglin and Ye, Shenglong and Shao, Jie and others},
  journal={arXiv preprint arXiv:2508.18265},
  year={2025}
}

@inproceedings{yang2025egolife,
  title={Egolife: Towards egocentric life assistant},
  author={Yang, Jingkang and Liu, Shuai and Guo, Hongming and Dong, Yuhao and Zhang, Xiamengwei and Zhang, Sicheng and Wang, Pengyun and Zhou, Zitang and Xie, Binzhu and Wang, Ziyue and others},
  booktitle={Proceedings of the Computer Vision and Pattern Recognition Conference},
  pages={28885--28900},
  year={2025}
}

@inproceedings{caba2015activitynet,
  title={Activitynet: A large-scale video benchmark for human activity understanding},
  author={Caba Heilbron, Fabian and Escorcia, Victor and Ghanem, Bernard and Carlos Niebles, Juan},
  booktitle={Proceedings of the ieee conference on computer vision and pattern recognition},
  pages={961--970},
  year={2015}
}

@inproceedings{deng2023large,
  title={A large-scale study of spatiotemporal representation learning with a new benchmark on action recognition},
  author={Deng, Andong and Yang, Taojiannan and Chen, Chen},
  booktitle={Proceedings of the IEEE/CVF International Conference on Computer Vision},
  pages={20519--20531},
  year={2023}
}

@inproceedings{takahashi2024abstractive,
  title={Abstractive multi-video captioning: Benchmark dataset construction and extensive evaluation},
  author={Takahashi, Rikito and Kiyomaru, Hirokazu and Chu, Chenhui and Kurohashi, Sadao},
  booktitle={Proceedings of the 2024 Joint International Conference on Computational Linguistics, Language Resources and Evaluation (LREC-COLING 2024)},
  pages={57--69},
  year={2024}
}

@inproceedings{li2024mvbench,
  title={Mvbench: A comprehensive multi-modal video understanding benchmark},
  author={Li, Kunchang and Wang, Yali and He, Yinan and Li, Yizhuo and Wang, Yi and Liu, Yi and Wang, Zun and Xu, Jilan and Chen, Guo and Luo, Ping and others},
  booktitle={Proceedings of the IEEE/CVF Conference on Computer Vision and Pattern Recognition},
  pages={22195--22206},
  year={2024}
}

@inproceedings{fu2025video,
  title={Video-mme: The first-ever comprehensive evaluation benchmark of multi-modal llms in video analysis},
  author={Fu, Chaoyou and Dai, Yuhan and Luo, Yongdong and Li, Lei and Ren, Shuhuai and Zhang, Renrui and Wang, Zihan and Zhou, Chenyu and Shen, Yunhang and Zhang, Mengdan and others},
  booktitle={Proceedings of the Computer Vision and Pattern Recognition Conference},
  pages={24108--24118},
  year={2025}
}

@article{hu2025video,
  title={Video-mmmu: Evaluating knowledge acquisition from multi-discipline professional videos},
  author={Hu, Kairui and Wu, Penghao and Pu, Fanyi and Xiao, Wang and Zhang, Yuanhan and Yue, Xiang and Li, Bo and Liu, Ziwei},
  journal={arXiv preprint arXiv:2501.13826},
  year={2025}
}

@article{ye2024gmai,
  title={Gmai-mmbench: A comprehensive multimodal evaluation benchmark towards general medical ai},
  author={Ye, Jin and Wang, Guoan and Li, Yanjun and Deng, Zhongying and Li, Wei and Li, Tianbin and Duan, Haodong and Huang, Ziyan and Su, Yanzhou and Wang, Benyou and others},
  journal={Advances in Neural Information Processing Systems},
  volume={37},
  pages={94327--94427},
  year={2024}
}

@article{sepehri2024mediconfusion,
  title={MediConfusion: Can you trust your AI radiologist? Probing the reliability of multimodal medical foundation models},
  author={Sepehri, Mohammad Shahab and Fabian, Zalan and Soltanolkotabi, Maryam and Soltanolkotabi, Mahdi},
  journal={arXiv preprint arXiv:2409.15477},
  year={2024}
}

@article{anderson2004eye,
  title={Eye movements do not reflect retrieval processes: Limits of the eye-mind hypothesis},
  author={Anderson, John R and Bothell, Dan and Douglass, Scott},
  journal={Psychological Science},
  volume={15},
  number={4},
  pages={225--231},
  year={2004},
  publisher={SAGE Publications Sage CA: Los Angeles, CA}
}

@article{ma2023eye,
  title={Eye-gaze-guided vision transformer for rectifying shortcut learning},
  author={Ma, Chong and Zhao, Lin and Chen, Yuzhong and Wang, Sheng and Guo, Lei and Zhang, Tuo and Shen, Dinggang and Jiang, Xi and Liu, Tianming},
  journal={IEEE Transactions on Medical Imaging},
  volume={42},
  number={11},
  pages={3384--3394},
  year={2023},
  publisher={IEEE}
}

@article{ozsoy2025egoexor,
  title={EgoExOR: An Ego-Exo-Centric Operating Room Dataset for Surgical Activity Understanding},
  author={{\"O}zsoy, Ege and Mamur, Arda and Tristram, Felix and Pellegrini, Chantal and Wysocki, Magdalena and Busam, Benjamin and Navab, Nassir},
  journal={arXiv preprint arXiv:2505.24287},
  year={2025}
}

\clearpage

\appendix
\begin{center}
     \Large \textbf{Appendix for MedGaze-Bench}
\end{center}
\textbf{Abstract.}
Appendix A outlines our four-stage prompt construction pipeline—clinical modeling, purpose specification, content constraints, and structural constraints—that generates first-person, gaze-aware, multimodal clinical questions across seven evaluation scenarios. Seven prompt templates are provided as examples.
\section{Details on Prompt Design and Templates}
\label{sec:appendix_prompts}
% System Prompt 和 User Prompt...
Our prompt construction pipeline follows a four-stage sequential process: clinical modeling → purpose specification → content constraints → structural constraints. This design ensures that the generated questions cognitively mirror the intention-driven decision-making of clinicians operating in real-world settings. In the clinical modeling stage, each prompt begins with a first-person narrative (“I am...”) to explicitly assign the model a specific clinical role such as attending surgeon, obstetrician, or radiology mentor and dynamically embeds high-fidelity contextual details of the current scenario. These include the precise surgical procedure (e.g., “open cholecystectomy”), the exact phase within the standardized operative protocol (SOP), the anatomical site, the number of available visual frames, a summary of expert eye-gaze patterns, team configuration (e.g., “anesthesiologist is present; scrub nurse is handing the suction device”), and relevant clinical guidelines (e.g., NICE 2025 or ACOG 2024). This contextual grounding shifts question generation away from generic medical knowledge recall and firmly anchors it to the specific visual evidence provided.

Following scene establishment, the prompt enters the purpose specification stage, which clearly defines the assessment objective for that template. We developed seven specialized prompt templates corresponding to seven distinct clinical evaluation scenarios: real-world open surgery; two phases of breech delivery simulation (pre-active waiting phase and active pushing phase); interpretation of chest X-rays and mammograms; and two robustness-focused tasks—Perceptual Reliability and Cognitive Reliability. Each template includes a tailored task description: for imaging interpretation prompts, the goal is to guide the model toward identifyingwhere critical evidence resides rather than stating a diagnosis; for Perceptual Reliability, the focus is on detecting whether the model erroneously selects items that are clinically plausible but absent from the visual frames; for Cognitive Reliability, the emphasis is on evaluating whether the model can recognize and reject false assumptions embedded in the question that contradict the visual evidence. To improve task fidelity, each template also provides a small set of compliant example utterances (e.g., “I have just called for the laparoscopic grasper—but where exactly should I place it?”) to illustrate how clinical intent should be translated into valid first-person questions.

In the content constraints stage, we enforce strict semantic and cognitive rules to ensure question quality and evaluative validity. The foremost requirement is mandatory use of the first-person “I” perspective: all questions must be phrased as “I need to…”, “I am unsure whether…”, or “I should check…”, avoiding third-person narration, passive voice, or abstract descriptions. This preserves the immediacy of the clinician’s in-the-moment cognitive state during procedural execution. Furthermore, each prompt must elicit exactly six questions that collectively span all six sub-dimensions of our three-dimensional clinical intention framework. Critically, we enforce multimodal dependency: especially in surgical and simulation contexts, the correct answer must rely on the joint interpretation of image content and the visual focus indicated by eye-gaze data. However, explicit coordinate references (e.g., x=0.6 or point 3) are strictly prohibited in the question stem; instead, natural-language spatial references such as “in the area I’m currently looking at” or “slightly left of center in my field of view” must be used. This design tests whether the model can effectively integrate visual attention with linguistic reasoning without exposing technical artifacts.

Finally, in the structural constraints stage, each prompt appends concise definitions of the six intention sub-dimensions along with illustrative question templates, and mandates that the output adhere to a standardized JSON format. This dual safeguard—conceptual guidance plus rigid output structure—ensures alignment with the intended evaluation dimensions while enabling reliable downstream parsing and automated scoring.

\subsection{Prompt for real-world open surgery}

% \begin{tcolorbox}[colframe=blue!50!black, colback=blue!10!white, coltitle=white, boxrule=0.8mm, width=0.5\textwidth, arc=4mm, auto breakable, title={Prompt for real-world open surgery}]
% \textbf{Scenario:}

% Real first-person operating room perspective (I am the operating surgeon)  
% \begin{itemize}
%     \item \textbf{Current phase:} \{surgical\_phase\}
%     \item \textbf{Procedure name:} \{procedure\_name\}
%     \item \textbf{Surgical site:} \{surgical\_site\}
%     \item \textbf{Attached frames:} \{frames\}
%     \item \textbf{My gaze:} \{gaze\}
% \end{itemize}

% \textbf{NON-NEGOTIABLE RULES:}
% \begin{enumerate}
%     \item Every question must be phrased strictly in first-person ``I'' form.
%     \item All five options must sound subjectively plausible to an experienced surgeon.
%     \item The correct answer must contradict pre-2023 procedural ``muscle memory.''
%     \item At least 5 questions must incorporate 2023–2025 guideline updates or rare but lethal intraoperative details.
%     \item At least 4 questions must have correct answers that critically depend on the gaze coordinates provided in ``My gaze.''
%     \item Do NOT include specific numerical gaze coordinates (e.g., x=0.5) in any question; use only abstract references like ``in the video.''
% \end{enumerate}

% \textbf{Use the provided 3×2 dimension framework without modification.}
% \end{tcolorbox}

\begin{itemize}
    \item \textbf{Your task}: Generate exactly 6 expert-killer multiple-choice questions.
    \item \textbf{Scenario}:
        \begin{itemize}
            \item Real first-person operating room perspective (I am the operating surgeon)
            \item Current phase: \texttt{\{surgical\_phase\}}
            \item Procedure name: \texttt{\{procedure\_name\}}
            \item Surgical site: \texttt{\{surgical\_site\}}
            \item Attached frames: \texttt{\{frames\}}
            \item My gaze: \texttt{\{gaze\}}
        \end{itemize}
    \item \textbf{NON-NEGOTIABLE RULES}:
        \begin{itemize}
            \item Every question must be phrased strictly in first-person ``I'' form.
            \item All five options must sound subjectively plausible to an experienced surgeon.
            \item The correct answer must contradict pre-2023 procedural ``muscle memory.''
            \item At least 5 questions must incorporate 2023--2025 guideline updates or rare but lethal intraoperative details.
            \item At least 4 questions must have correct answers that critically depend on the gaze coordinates provided in ``My gaze.''
            \item Do NOT include specific numerical gaze coordinates (e.g., \(x = 0.5\)) in any question; use only abstract references like ``in the video.''
        \end{itemize}
    \item Use the provided 3$\times$2 dimension framework without modification.
\end{itemize}

\subsection{Prompt for breech delivery}
% \begin{tcolorbox}[colframe=blue!50!black, colback=blue!10!white, coltitle=white, boxrule=0.8mm, width=0.5\textwidth, arc=4mm, auto breakable, title={Prompt for Breech Delivery}]

% \textbf{Scenario:}

% Real first-person operating room perspective (I am the operating surgeon)
% \begin{itemize}
% \item \textbf{Current phase:} \{surgical\_phase\}
% \item \textbf{Procedure name:} \{procedure\_name\}
% \item \textbf{Surgical site:} \{surgical\_site\}
% \item \textbf{Attached frames:} \{frames\}
% \item \textbf{My gaze:} \{gaze\}
% \end{itemize}

% \textbf{NON-NEGOTIABLE RULES:}
% \begin{enumerate}
% \item Every question must be phrased strictly in first-person `I'' form.
%     \item All five options must sound subjectively plausible to an experienced obstetrician.
%     \item The correct answer must contradict pre-2023 procedural `muscle memory.''
% \item At least 5 questions must be based on 2023–2025 guideline updates or rare but lethal intraoperative details.
% \item At least 4 questions must have correct answers that critically depend on the gaze coordinates provided in `My gaze.''
%     \item Do NOT include specific numerical gaze coordinates (e.g., x=0.5) in any question; use only abstract references like `in the video.''
% \end{enumerate}

% \textbf{Use the provided 3×2 dimension framework without modification.}
% \end{tcolorbox}

\begin{itemize}
    \item \textbf{Your task}: Generate exactly 6 “expert-killer” single-best-answer questions.
    \item \textbf{Scenario}:
        \begin{itemize}
            \item Real first-person operating room perspective (I am the operating surgeon)
            \item Current phase: \texttt{\{delivery\_phase\}}
            \item Procedure name: \texttt{\{procedure\_name\}}
            \item Surgical site: \texttt{\{surgical\_site\}}
            \item Attached frames: \texttt{\{frames\}}
            \item My gaze: \texttt{\{gaze\}}
        \end{itemize}
    \item \textbf{Your hands are on the fetus. You must master these four lethal critical skills:}
        \begin{itemize}
            \item Precise timing and dosing of oxytocin (Sintocinon) augmentation.
            \item Accurate recognition of true active labor when the fetal scapulae become visible.
            \item Controlled delivery of extended arms using only maternal effort and gravity—never applying traction.
            \item Perfect execution of the Bracht maneuver without exerting any traction on the fetus.
        \end{itemize}
    \item \textbf{NON-NEGOTIABLE RULES}:
        \begin{itemize}
            \item Every question must be phrased strictly in first-person ``I'' form.
            \item All five options must sound subjectively plausible to an experienced obstetrician.
            \item The correct answer must contradict pre-2023 procedural ``muscle memory.''
            \item At least 5 questions must be based on 2023--2025 guideline updates or rare but lethal intraoperative details.
            \item At least 4 questions must have correct answers that critically depend on the gaze coordinates provided in ``My gaze.''
            \item Do NOT include specific numerical gaze coordinates (e.g., \(x = 0.5\)) in any question; use only abstract references such as ``in the video.''
        \end{itemize}
    \item Use the provided 3$\times$2 dimension framework without modification.
\end{itemize}

\begin{itemize}
    \item \textbf{Your task}: You are generating a gaze-dependent VideoQA (video question-answering) benchmark dataset.
    \item \textbf{Scenario}:
        \begin{itemize}
            \item Real first-person breast radiologist perspective (I am the radiologist)
            \item View: \texttt{\{view\_position\_full\}} 
            \item Short Findings: \texttt{\{findings\}}
            \item BI-RADS Category: \texttt{\{birads\}}
            \item Gaze Pattern: \texttt{\{gaze\}}
        \end{itemize}
    \item \textbf{Spatial Coordinate Reference (Mental Model)}:
        \begin{itemize}
            \item Image Orientation: Standard DICOM format. (0,0) is top-left.
            \item Pectoral Muscle: Superior/axillary side (x 0.7, depending on laterality).
            \item Nipple: Anterior edge (center Y-axis, extreme X-axis).
        \end{itemize}
    \item \textbf{NON-NEGOTIABLE RULES}:
        \begin{itemize}
            \item Every question must be phrased strictly in first-person ``I'' form.
            \item All five options must sound subjectively plausible to an experienced breast radiologist.
            \item The correct answers to at least 4 questions must critically depend on the gaze coordinates provided in ``My gaze.''
            \item Infer the lesion’s underlying nature based on BI-RADS and gaze: combine the BI-RADS category (2–5) with gaze dwell time to reasonably assume malignant features, benign features, or calcification distribution patterns, thereby constructing a hidden but consistent “ground-truth pathology.”
            \item Design blind-test questions that are unanswerable without gaze data: distractors must be generally plausible, but only resolvable using the specific visual attention pattern (gaze) provided.
            \item Strictly adhere to anatomical constraints of 2D mammography:
                \begin{itemize}
                    \item In CC view, avoid ``upper vs. lower'' descriptors (anatomically indistinguishable); valid terms include lateral/medial, retroareolar, deep/posterior.
                    \item In MLO view, use superior/inferior and axillary tail; avoid overly precise medial/lateral distinctions.
                \end{itemize}
        \end{itemize}
        \item Use the provided 3$\times$2 dimension framework without modification.
\end{itemize}

\subsection{Prompt for Mammograph interpretation}
\begin{itemize}
    \item \textbf{Your task}: Generate 6 questions strictly in the form of:“Which area should I observe?” or “Where is the evidence located?”
    \item \textbf{Scenario}:
        \begin{itemize}
            \item Current diagnosis: \texttt{\{diagnosis\}}
            \item My gaze focus: \texttt{\{gaze\}}
        \end{itemize}
    \item \textbf{Non-negotiable rules}:  
        \begin{itemize}
            \item Every question must be phrased strictly in first-person ``I'' form.
            \item All five options must sound subjectively plausible to an experienced radiologist.
            \item Do not mention specific gaze coordinates or point indices (e.g., x=0.5, points 2, 3, 5, 6). Use only abstract references such as ``the area I’m currently fixating on" or ``the region in my field of view".
            \item The correct answers to at least 4 questions must critically depend on the gaze coordinates provided in ``My gaze".
            \item Questions may describe image content and current intent, but must not directly or indirectly reveal the lesion location (e.g., “small left pleural effusion” should only be phrased as ``small pleural effusion").
            \item Options may include visually or clinically similar distractors to test differential diagnostic reasoning.
        \end{itemize}

    \item Example questions:
        \begin{itemize}
            \item Where is the evidence located in this image?
            \item Which area should I focus on to observe the critical signs of this diagnosis?
            \item Where should I look to confirm the presence of the lesion?
            \item In which region am I observing the most relevant clinical feature for this case?
            \item What part of the image am I fixating on to rule out a potential finding?
            \item Which region in my field of view should I analyze to determine the lesion's characteristics?
        \end{itemize}
        \item Use the provided 3$\times$2 dimension framework without modification.
\end{itemize}

\subsection{Prompt for Perceptual Reliability}
\begin{itemize}
    \item \textbf{Your task}: Generate exactly 6 “expert-killer” multiple-choice questions designed to expose models that blindly accept false premises in the prompt—even when visual evidence clearly contradicts them.
    \item \textbf{Scenario}:
        \begin{itemize}
            \item Real first-person operating room perspective (I am the primary surgeon)
            \item Current phase: \texttt{\{sop\_phase\}}
            \item Procedure name: \texttt{\{procedure\_name\}}
            \item Surgical site: \texttt{\{surgical\_site\}}
            \item Number of attached frames: \texttt{\{frames\}}
            \item My gaze focus: \texttt{\{gaze\}}
        \end{itemize}
    \item \textbf{NON-NEGOTIABLE HARD RULES}:
        \begin{itemize}
            \item Every question must:
                \begin{itemize}
                    \item Be phrased strictly in first-person "I" form.
                    \item Contain a false or unsupported assumption in the stem that directly contradicts the visual evidence (e.g., asserting the presence of a non-existent team member, action, instrument, or guideline). (Critically important)
                    \item Align closely with the core intent of its assigned subtype.
                    \item Embed the question within a clinically plausible context that includes the procedure name, current phase, and anatomical location, leading naturally to the query.
                \end{itemize}
            \item Regarding the five options:
                \begin{itemize}
                    \item The single correct answer must explicitly reject or correct the false premise. (Critically important)
                    \item The other four distractors must endorse the false assumption, offering responses that sound reasonable but are factually wrong. (Critically important)
                    \item All five options must appear subjectively credible to an experienced surgeon.
                    \item Each option must include a brief explanatory phrase, intent clarification, or descriptive justification beyond just the answer choice.
                \end{itemize}
            \item The correct answer must be verifiable solely from visible content—no inference beyond direct observation is allowed.
            \item All 6 questions must require attention to fine-grained visual details consistent with “My gaze” (e.g., instrument tip, tissue color, hand position).
            \item Do not reference specific gaze coordinates (e.g., \(x = 0.5\)); use only abstract phrases like "the area I’m looking at" or "what’s currently in my field of view".
        \end{itemize}
    \item Use the provided 3×2 dimension framework without modification.
\end{itemize}

\subsection{Prompt for Cognitive Reliability}
\begin{itemize}
    \item \textbf{Your task}: Generate exactly 6 VQA-style multiple-choice questions—one for each subtype—specifically designed to expose perception-fact conflict hallucinations, where models incorrectly select items that are not actually visible in the provided visual frames.
    \item \textbf{Scenario}:
        \begin{itemize}
            \item Real first-person operating room perspective (I am the primary surgeon)
            \item Current phase: \texttt{\{sop\_phase\}}
            \item Procedure name: \texttt{\{procedure\_name\}}
            \item Surgical site: \texttt{\{surgical\_site\}}
            \item Number of attached frames: \texttt{\{frames\}}
            \item My gaze focus: \texttt{\{gaze\}}
        \end{itemize}
    \item \textbf{NON-NEGOTIABLE HARD RULES}:
        \begin{itemize}
            \item Every question must:
                \begin{itemize}
                    \item Be phrased strictly in first-person ``I'' form.
                    \item Ask about something actually visible, currently in use, or directly observable in the attached frames.
                    \item Align precisely with the core intent of its assigned subtype.
                    \item Embed the query within a clinically plausible context that includes the procedure name, current surgical phase, and anatomical location, leading naturally to the question.
                \end{itemize}
            \item Regarding the five options:
                \begin{itemize}
                    \item Exactly one option must correspond to an object, instrument, tissue, or action visibly present in the frames.
                    \item The other four options must be highly relevant and common for this surgical phase—but absent from all provided frames.
                    \item All five options must appear subjectively reasonable and credible to an experienced surgeon.
                    \item Each option must include a brief explanatory phrase, rationale, or descriptive justification beyond the answer itself.
                \end{itemize}
            \item The correct answer must be verifiable solely through direct visual observation—no inference beyond what is shown is permitted.
            \item All 6 questions must require attention to fine-grained visual details consistent with ``My gaze'' (e.g., instrument tip, tissue color, hand position).
            \item Do not reference specific gaze coordinates (e.g., \(x = 0.5\)); use only abstract phrases like ``the area I’m looking at'' or ``what’s currently in my field of view.''
        \end{itemize}
    \item Use the provided 3×2 dimension framework without modification.
\end{itemize}
\subsection{Provided 6 dimension framework}
\begin{itemize}
    \item \textbf{Clinical Spatial Intent} (Where is the surgeon looking?):
        \begin{itemize}
            \item 1.1: Relative Positioning (Relative positioning between objects)
            \item 1.2: Global Positioning (Global position within the image)
        \end{itemize}
    \item \textbf{Clinical Temporal Intent} (When is the surgeon looking?):
        \begin{itemize}
            \item 2.1: Temporal Intent (What will be done next?)
            \item 2.2: Causal Intent (What led me to do this?)
        \end{itemize}
    \item \textbf{Clinical Standard Intent} (Is the surgeon looking according to standard protocols?):
        \begin{itemize}
            \item 3.1: Critical Checkpoint (Short-term continuous monitoring for adherence to clinical surgical guidelines)
            \item 3.2: Continuous Watching (Long-term discrete monitoring for adherence to clinical surgical guidelines)
        \end{itemize}
\end{itemize}

% 建议 2: Case Study 
% \section{Additional Qualitative Results and Case Studies}
% \label{sec:appendix_cases}

\smallskip
\smallskip
\smallskip

\end{document}